\documentclass[fleqn,10pt]{wlscirep}
\usepackage[utf8]{inputenc}
\usepackage[T1]{fontenc}
\usepackage{amsmath,amssymb}
\usepackage{graphicx}
\usepackage{booktabs,multirow,array,tabularx,makecell}
\usepackage{threeparttable}
\usepackage{colortbl}
\usepackage{xcolor}
\usepackage{pifont}
\usepackage{inconsolata} 
\usepackage{pdflscape}
\usepackage{float,newfloat}
\usepackage{enumitem}
\usepackage{setspace}
\usepackage{textcomp}
\usepackage{microtype}
\usepackage{caption,subcaption}

\newcolumntype{C}[1]{>{\centering\arraybackslash}p{#1}}
\newcolumntype{R}[1]{>{\raggedleft\arraybackslash}p{#1}}
\newcolumntype{L}[1]{>{\raggedright\arraybackslash}p{#1}}

\DeclareMathAlphabet{\mathcal}{OMS}{cmsy}{m}{n}

\DeclareFloatingEnvironment[name={Supplementary Figure}]{suppfigure}
\DeclareFloatingEnvironment[name={Supplementary Table}]{supptable}

\title{TriALS: Triphasic-Aided Liver Lesion Segmentation Benchmark in Non-Contrast CT}

\author[1]{Marawan Elbatel}

\author[2,3]{Mohamed Ghonim}
\author[4]{Jiaji Mao}
\author[5]{Zhuosheng Lin}

\author[6,7]{Katharina Eckstein}
\author[6,7]{Andrés Martínez Mora}
\author[6,8]{Jonathan Deissler}
\author[6,8]{Maximilian Rokuss}
\author[6,7]{Constantin Ulrich}

\author[9]{Zdravko Marinov}
\author[4]{Wenhui Deng}
\author[4]{Baoxun Li}
\author[4]{Huijun Hu}
\author[4]{Jun Shen}

\author[2,3]{Mohanad Ghonim}

\author[2,3]{Khadiga Omar Nassar}
\author[2,3]{Mariam Elbakry}
\author[2]{Menna Dyab}
\author[2,3]{Amr Muhammad Abdo Salem}
\author[2,3]{Nouran Elghitany}
\author[2,3]{Noha Elghitany}
\author[1]{Yi Qin}
\author[1]{Xuanqi Huang}
\author[1]{Haonan Wang}

\author[10]{Shao-Woo Yen}

\author[11]{Ahmed Elghamry Saba}
\author[12]{Salma Ahmad}

\author[13]{Xinyan Fang}
\author[13]{Jiahao Zhang}
\author[13]{Xiaodi Wang}
\author[13]{Xinghua Ma}
\author[13]{Gongning Luo}

\author[14]{Jessica C. Delmoral}
\author[14]{João Manuel R.S. Tavares}

\author[15]{Ankan Deria}
\author[15]{Adinath Dukre}
\author[15]{Yutong Xie}
\author[15]{Imran Razzak}

\author[16]{Dongwook Kim}
\author[16]{Matthew Choi}

\author[17]{Hanxiao Zhang}
\author[17]{Minghui Zhang}
\author[17]{Xin You}

\author[18]{Abdul Qayyum}
\author[18]{Steven A. Niederer}
\author[19]{Moona Mazher}

\author[20]{Rachika E. Hamadache}
\author[20]{Ricardo Montoya-del-Angel}
\author[20]{Robert Martí}
\author[20]{Xavier Lladó}

\author[21]{Toufiq Musah}
\author[21]{Livingstone Eli Ayivor}

\author[22]{Enrique Almar-Munoz}
\author[22]{Agnes Mayr}

\author[23]{Kaouther Mouheb}
\author[23]{Esther E. Bron}
\author[23]{Stefan Klein}

\author[3]{Ahmed Abouelhoda}
\author[3]{Amira Adel}
\author[3]{Susan Adil Ali}
\author[9]{Rainer Stiefelhagen}
\author[6,7,8,24,25]{Klaus H. Maier-Hein}
\author[6,24]{Fabian Isensee}
\author[2,3]{Aya Yassin}
\author[1]{Xiaomeng Li}

\affil[1]{Department of Electronic and Computer Engineering, The Hong Kong University of Science and Technology, Hong Kong SAR}
\affil[2]{AI Center of Excellence, Ain Shams University, Cairo, Egypt}
\affil[3]{Department of Radiology, Ain Shams University, Cairo, Egypt}
\affil[4]{Department of Radiology, Guangdong Provincial Key Laboratory of Malignant Tumor Epigenetics and Gene Regulation, Sun Yat-Sen Memorial Hospital, Sun Yat-Sen University, Guangzhou, China}
\affil[5]{Nanfang Hospital, Southern Medical University, Guangzhou, China}
\affil[6]{Division of Medical Image Computing, German Cancer Research Center (DKFZ), Heidelberg, Germany}
\affil[7]{Medical Faculty Heidelberg, Heidelberg University, Germany}
\affil[8]{Faculty of Mathematics and Computer Science, Heidelberg University, Germany}
\affil[9]{Karlsruhe Institute of Technology, Germany}
\affil[10]{Institute of Biomedical Engineering, National Yang Ming Chiao Tung University, Taiwan}
\affil[11]{NewGiza University, Giza, Egypt}
\affil[12]{Department of Computer Engineering and Data Science, Ain Shams University, Cairo, Egypt}
\affil[13]{Faculty of Computing, Harbin Institute of Technology, Harbin, China}
\affil[14]{Faculdade de Engenharia, Universidade do Porto, Porto, Portugal}
\affil[15]{Mohamed Bin Zayed University of Artificial Intelligence, Abu Dhabi, UAE}
\affil[16]{Korea University, Seoul, South Korea}
\affil[17]{Shanghai Key Laboratory of Flexible Medical Robotics, Tongren Hospital, Institute of Medical Robotics, Shanghai Jiao Tong University, Shanghai, China}
\affil[18]{National Heart and Lung Institute, Faculty of Medicine, Imperial College London, London, UK}
\affil[19]{Hawkes Institute, Department of Computer Science, University College London, London, UK}
\affil[20]{Computer Vision and Robotics Institute (ViCOROB), University of Girona, Girona, Spain}
\affil[21]{Department of Computer Engineering, Kwame Nkrumah University of Science and Technology, Ghana}
\affil[22]{Department of Radiology, Medical University of Innsbruck, Innsbruck, Austria}
\affil[23]{Erasmus MC, Rotterdam, The Netherlands}
\affil[24]{Helmholtz Imaging, DKFZ, Heidelberg, Germany}
\affil[25]{Pattern Analysis and Learning Group, Dept. of Radiation Oncology, Heidelberg University Hospital, Germany}

\begin{document}


\maketitle
\clearpage

\section*{ABSTRACT}
Automated segmentation of liver lesions on non-contrast computed tomography (NCCT) is clinically important but fundamentally challenging, particularly in low-resource settings across Africa and Asia where contrast agents are frequently unavailable. Progress has been limited by the absence of annotated NCCT benchmarks. Here we describe the TriALS challenge for automated liver lesion segmentation under contrast-limited conditions, supported by a multi-centre dataset of 150 cases with four-phase CT acquisitions (600 volumes) from Egyptian and Chinese institutions. Algorithms were evaluated on 70 cases from three institutions, including an independent external cohort. The top-performing method achieved a mean venous-phase Dice of 0.754, consistent with human-level performance, yet dropped to 0.57 on NCCT. On external validation, the leading method outperformed off-the-shelf models by up to 28\% in Dice on NCCT. Algorithm performance was most strongly predicted by training data scale and pre-training strategy. A cross-year comparison exposed a persistent perceptual barrier on NCCT that scaling pre-training alone cannot overcome. Data, annotations, and code are available at \url{https://github.com/xmed-lab/TriALS}.


\section*{Introduction}
Liver cancer is a leading cause of cancer death in developing nations, with Egypt bearing one of the highest global burdens~\cite{liver_cancer_risk_intro,liver_stats_in_egypt_intro}. Accurate identification and delineation of hepatic lesions guides critical therapeutic decisions, including resection, ablation, and transarterial treatment. Contrast-enhanced computed tomography (CT) is the diagnostic standard for lesion detection and staging. However, iodinated contrast agents are frequently unavailable owing to supply-chain disruptions that disproportionately affect low-resource hospitals~\cite{contrast_supply_issue}. In Egypt, recurring shortages have severely curtailed routine multi-phase liver imaging~\cite{EgyptIndependent2025_CTContrastShortage}. Beyond supply constraints, contrast administration is contraindicated in patients with renal impairment or hypersensitivity. Non-contrast CT (NCCT) is consequently the default modality accessible in many resource-constrained clinical settings. On NCCT, lesion conspicuity is minimal because density differences between tumour and surrounding parenchyma are small. Radiologists therefore face a fundamentally harder diagnostic task when contrast is unavailable.

Artificial intelligence could enable reliable lesion detection on NCCT, yet progress is hindered by three compounding gaps. First, no multi-centre benchmark specifically designed for NCCT liver lesion segmentation evaluation exists. Existing public benchmarks such as LiTS~\cite{Lits_benchmark}, KiTS~\cite{heller2023kits21challengeautomaticsegmentation}, and AMOS~\cite{ji2022amoslargescaleabdominalmultiorgan} provide only single-phase contrast-enhanced images from single or well-resourced centres. Second, recent multi-phase datasets such as MCT-LTDiag~\cite{Wu2025} originate from single, well-resourced institutions. These cohorts target differential diagnosis rather than detection under contrast-limited conditions. Third, algorithmic advancements remain untested on the patient demographics and scanner configurations of the low-resource settings that most need them, particularly across African populations. Together, these gaps mean that no large-scale, non-contrast segmentation benchmark currently exists. Without such a resource, developing equitable diagnostic tools for contrast-limited healthcare systems remains impractical.

Here we present TriALS (Triphasic-Aided Liver Lesion Segmentation), to our knowledge the first multi-centre benchmark for liver lesion segmentation in non-contrast CT. The dataset comprises 150 cases (600 volumes across four phases) with aligned triphasic acquisitions from Egyptian and Chinese institutions, all annotations independently verified by board-certified radiologists. 

TriALS was organised across two consecutive editions of the Medical Image Computing and Computer-Assisted Intervention (MICCAI) conference to benchmark liver lesion segmentation in non-contrast computed tomography (CT). The 2024 edition used data exclusively from Ain Shams University, Egypt, attracting six teams. The 2025 edition introduced Chinese data from Sun Yat-Sen Memorial Hospital and an independent external test cohort, drawing eight teams across three continents (Europe, Asia, and Africa). Across both editions, we evaluated 13 unique teams on test sets spanning three centres, including an independent external cohort from Nanfang Hospital, Southern Medical University (Guangzhou, China). By quantifying human performance ceilings through inter-reader agreement, we show that the top method achieves a mean venous-phase Dice of 0.754 across all three test centres, consistent with human-level performance, and outperforms contemporary off-the-shelf models such as VoxTell~\cite{rokuss2025voxtell} and PASTA~\cite{lei2026syntheticdatadrivenradiologyfoundation_pastas} by up to +28\% absolute Dice on NCCT. A cross-year comparison reveals a +4.2\% absolute Dice improvement from 2024 to 2025 when evaluating against the full lesion burden derived from all contrast phases, yet no improvement when evaluating only against lesions visible on NCCT alone, exposing a fundamental perceptual barrier that scaling pre-training data alone cannot overcome. TriALS represents an initial step toward broader regional imaging initiatives, including pan-African and pan-Asian collaborations, to map the diverse lesion burden across underrepresented populations.

\section*{Results}

\subsection*{Challenge design and cohort overview}

\begin{figure}[htbp]
  \centering
  \includegraphics[width=\linewidth]{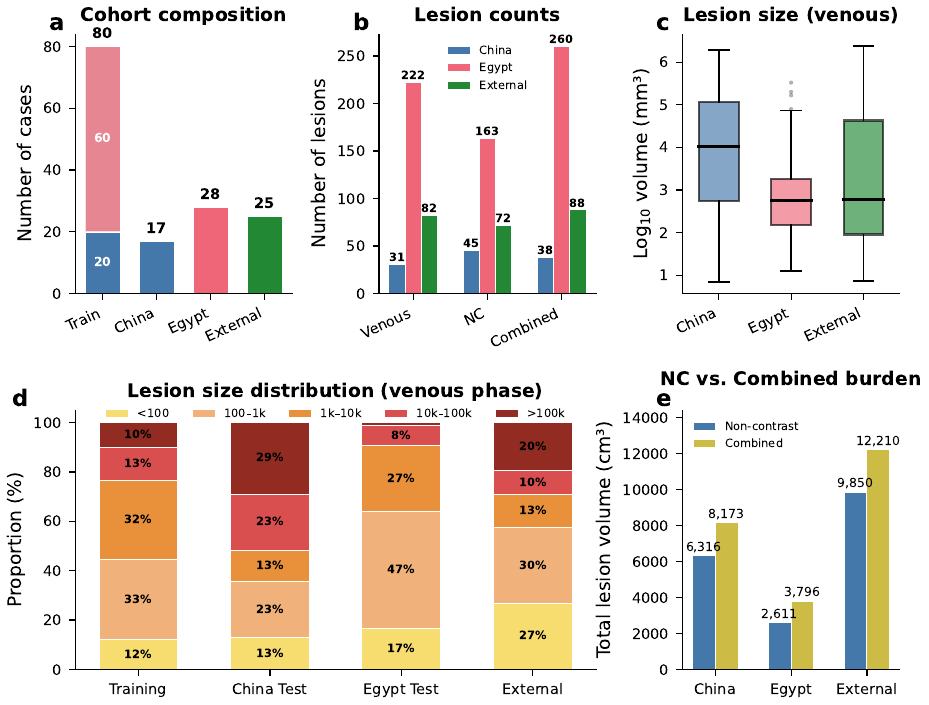}
  \caption{\textbf{TriALS cohort design and dataset characteristics.}
  \textbf{a,} Number of cases per split; training bars subdivided by country (China/Egypt).
  \textbf{b,} Lesion counts per centre. Per-phase lesion counts are computed as the number of connected components within each phase's mask. In cohorts with clustered lesions (e.g., China), contiguous coverage across contrast phases can merge adjacent non-contrast components into a single connected component in the fused mask, yielding combined counts that may be lower than non-contrast counts despite greater combined lesion burden (panel e).
  \textbf{c,} Venous-phase lesion volume distributions (log$_{10}$\,mm$^3$).
  \textbf{d,} Proportional lesion size distribution per cohort.
  \textbf{e,} Non-contrast versus combined total lesion volume (cm$^3$) per testing centre, quantifying the lesion burden captured by each annotation paradigm. Combined annotations capture greater lesion burden than non-contrast annotations in all three cohorts (China: +29\%; Egypt: +45\%; External: +24\%), confirming that a substantial fraction of hepatic lesions remains occult without contrast enhancement.}
  \label{fig:cohort}
\end{figure}

The complete dataset comprises 150 cases with four-phase CT acquisitions (non-contrast, arterial, portal venous, delayed) drawn from three institutions~(Fig.~\ref{fig:cohort}; Table~\ref{tab:cohort}). Training data included 80 cases (20~China, 60~Egypt), and three independent test sets were assembled: China ($n{=}17$), Egypt ($n{=}28$), and an external cohort ($n{=}25$) from Nanfang Hospital, Southern Medical University. This structure enabled evaluation of generalisation across institutions and populations, though we note that the external cohort shares demographic and scanner characteristics with the Chinese training data (see Limitations).
 
Lesion heterogeneity was substantial across cohorts. Egyptian lesions were predominantly small (median 556\,mm$^3$, IQR 151--1{,}813\,mm$^3$), with 64\% below 1{,}000\,mm$^3$ and  17\% below 100\,mm$^3$~(Table~\ref{tab:cohort}). Chinese lesions were markedly larger (median 10{,}658\,mm$^3$, IQR 554--117{,}859\,mm$^3$), with 29\% exceeding 100{,}000\,mm$^3$. The external cohort showed an intermediate profile (median 604\,mm$^3$, IQR 90--42{,}752\,mm$^3$)~(Supplementary Fig.~\ref{edfig:lesion_size}). This size disparity enables evaluation of algorithm robustness to markedly different lesion burden profiles within a single benchmark.

Annotations derived from multi-phase label fusion (hereafter, combined annotations) captured substantially greater total lesion burden than non-contrast-only annotations in all three cohorts (Egypt: 3,796 cm³ vs. 2,611 cm³, +45\%; External: 12,210 cm³ vs. 9,850 cm³, +24\%; China: 8,173 cm³ vs. 6,316 cm³, +29\%; Fig. 1e), confirming that a substantial fraction of hepatic lesions remains occult without contrast enhancement~(Fig.~\ref{fig:cohort}e). CT acquisition parameters also varied across institutions: in-plane spacing ranged from 0.686$\pm$0.072\,mm (external) to 0.862$\pm$0.095\,mm (Egypt), slice thickness from 0.86\,mm (Egypt) to 1.02\,mm (external), and volume depth from 240$\pm$69 slices (external) to 578$\pm$137 slices (Egypt)~(Table~\ref{tab:cohort}; Supplementary Fig.~\ref{fig:spacing}). Pairwise comparisons confirmed significant differences in acquisition parameters across all centre pairs (Mann--Whitney $U$, $p < 0.05$ for 8 of 9 comparisons; Table~\ref{tab:centre_pvalues}).

\subsection*{Inter-reader agreement establishes human performance ceilings}
 
\begin{figure}[htbp]
  \centering
  \includegraphics[width=\linewidth]{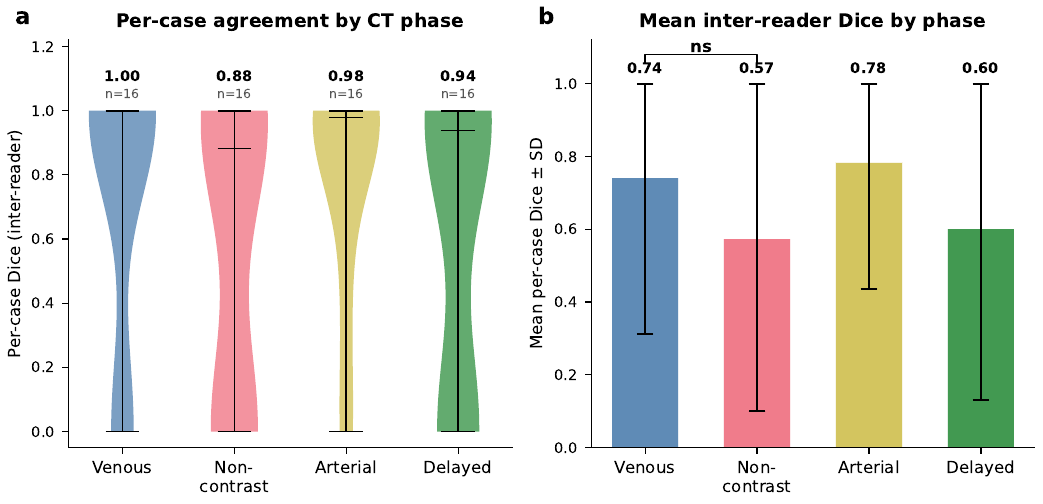}\\[0.5em]
  \includegraphics[width=\linewidth]{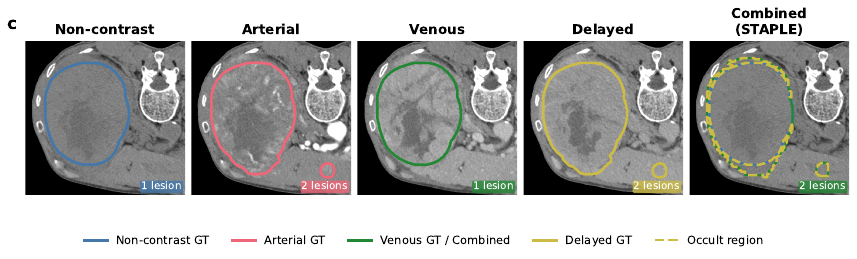}
  \caption{\textbf{Inter-reader agreement between two radiologists.}
    \textbf{a,} Per-case Dice violin distributions by CT phase ($n=16$ per phase); annotated values above each violin are medians (venous 1.00, non-contrast 0.88, arterial 0.98, delayed 0.94). The bimodal pattern reflects cases of complete agreement (DSC~$\approx$~1) versus complete disagreement on lesion visibility (DSC~$=$~0), which drives the gap between median and mean.
    \textbf{b,} Mean per-case Dice $\pm$\,SD per phase (venous 0.742, non-contrast 0.574, arterial 0.784, delayed 0.601) with Mann--Whitney significance annotation between venous and non-contrast.
    \textbf{c,} Same patient (External case~38) shown across all four CT phases at the same physical axial location (Z~$\approx$~647\,mm). Non-contrast CT reveals only 1 lesion (blue); the arterial phase reveals 2 lesions (red, arrows indicate the small occult lesion); venous (green) shows 1 lesion; delayed (yellow) shows 2 lesions. The combined annotation (rightmost, STAPLE fusion of all phases) captures 2 lesions on NCCT, with the occult region (yellow dashed, arrow) indicating a lesion detected only via contrast-phase fusion. This demonstrates how the multi-phase annotation paradigm captures the full lesion burden that single-phase NCCT reading misses.}
  \label{fig:interrater}
\end{figure}

Before evaluating algorithmic performance, we establish the human baseline. Two abdominal radiologists independently delineated liver lesions across all four CT phases~(Fig.~\ref{fig:interrater}). Inter-reader per-case Dice on venous-phase CT was 0.742 (SD~0.443, $n = 16$), with lesion-level precision 0.833 and recall 0.946~(Fig.~\ref{fig:interrater}a). Agreement was highest in the arterial phase (Dice 0.784) and declined to 0.601 in the delayed phase.

On non-contrast CT, inter-reader Dice fell to 0.574 (SD~0.490, $n = 16$), with precision dropping to 0.333 and recall to 0.769, a 16.8\% absolute Dice gap relative to venous-phase agreement~(Fig.~\ref{fig:interrater}b). The low non-contrast precision reflects a high false-discovery rate: two-thirds of lesions identified by one reader were missed by the other on non-contrast CT~(Fig.~\ref{fig:interrater}c). This directly motivates both the combined annotation paradigm, in which contrast-phase delineations guide non-contrast labels, and AI-assisted workflows.

The high standard deviations reflect a bimodal agreement pattern: radiologists either agreed perfectly on lesion delineation (DSC~$\approx$~1) or disagreed completely on lesion visibility (DSC~$=$~0). The top TriALS method (MIC) achieved a venous-phase Dice of 0.791 on the Egyptian cohort, exceeding the inter-reader mean by 4.9\%. Bootstrap resampling ($N = 10{,}000$) yielded a 95\% confidence interval of [0.540, 0.934] for the human ceiling; MIC falls within this interval, consistent with human-level performance on venous CT. The wide interval reflects the bimodal nature of inter-reader agreement and the small sample ($n = 16$), and should be interpreted as an approximate reference range rather than a precise threshold. For non-contrast CT, the best combined Dice on the external set (0.651) exceeded the inter-reader ceiling of 0.574. Performance on the harder Egyptian cohort (0.471) remained below the ceiling, indicating that the non-contrast challenge is not yet fully resolved.

\subsection*{Segmentation performance and ranking stability}
 
\begin{figure}[htbp]
  \centering
  \includegraphics[width=\linewidth]{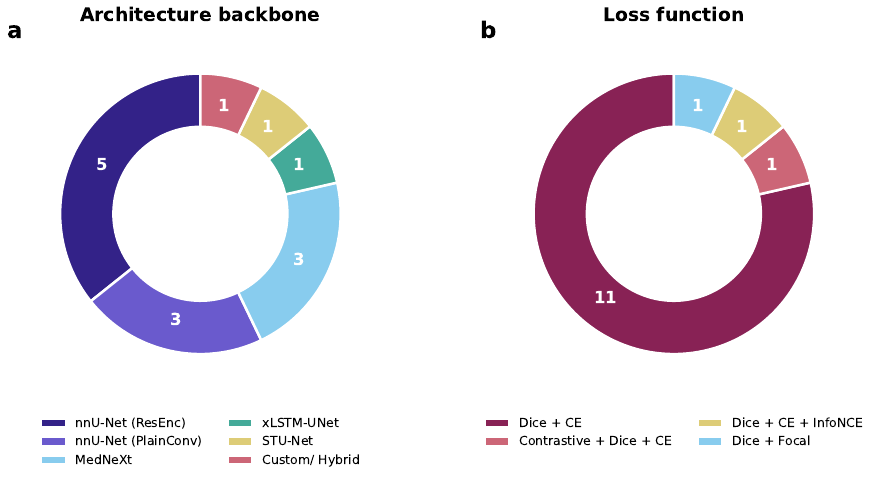}
  \caption{\textbf{Methodological landscape of participating teams.}
    \textbf{a,}~Architecture backbone distribution: nnU-Net variants (ResEnc and PlainConv) dominated, with MedNeXt, xLSTM-UNet, STU-Net, and custom hybrids also represented.
    \textbf{b,}~Loss function distribution: Dice + cross-entropy was used by 11 of 14 submissions; alternatives included contrastive, InfoNCE, and focal losses.}
  \label{edfig:method_overview}
\end{figure}

Across both challenge editions, 13 unique teams submitted 14 solutions~(Fig.~\ref{edfig:method_overview}). nnU-Net~\cite{isensee2021nnunet} served as the backbone for 12 of 14 submissions, with Dice + cross-entropy as the dominant loss function (11 of 14). The leading method employed large-scale multi-dataset pre-training across 86 public datasets via MultiTalent~\cite{ulrich2023multitalent}, combined with liver-ROI cropping and bootstrap-based model selection (Supplementary Note~1). Although nnU-Net provided the common backbone, several teams introduced architectural or training innovations, including custom multi-scale encoders and contrastive learning heads for small-lesion discrimination (Supplementary Note~1).
 
\begin{figure}[htbp]
  \centering
  \includegraphics[width=\linewidth]{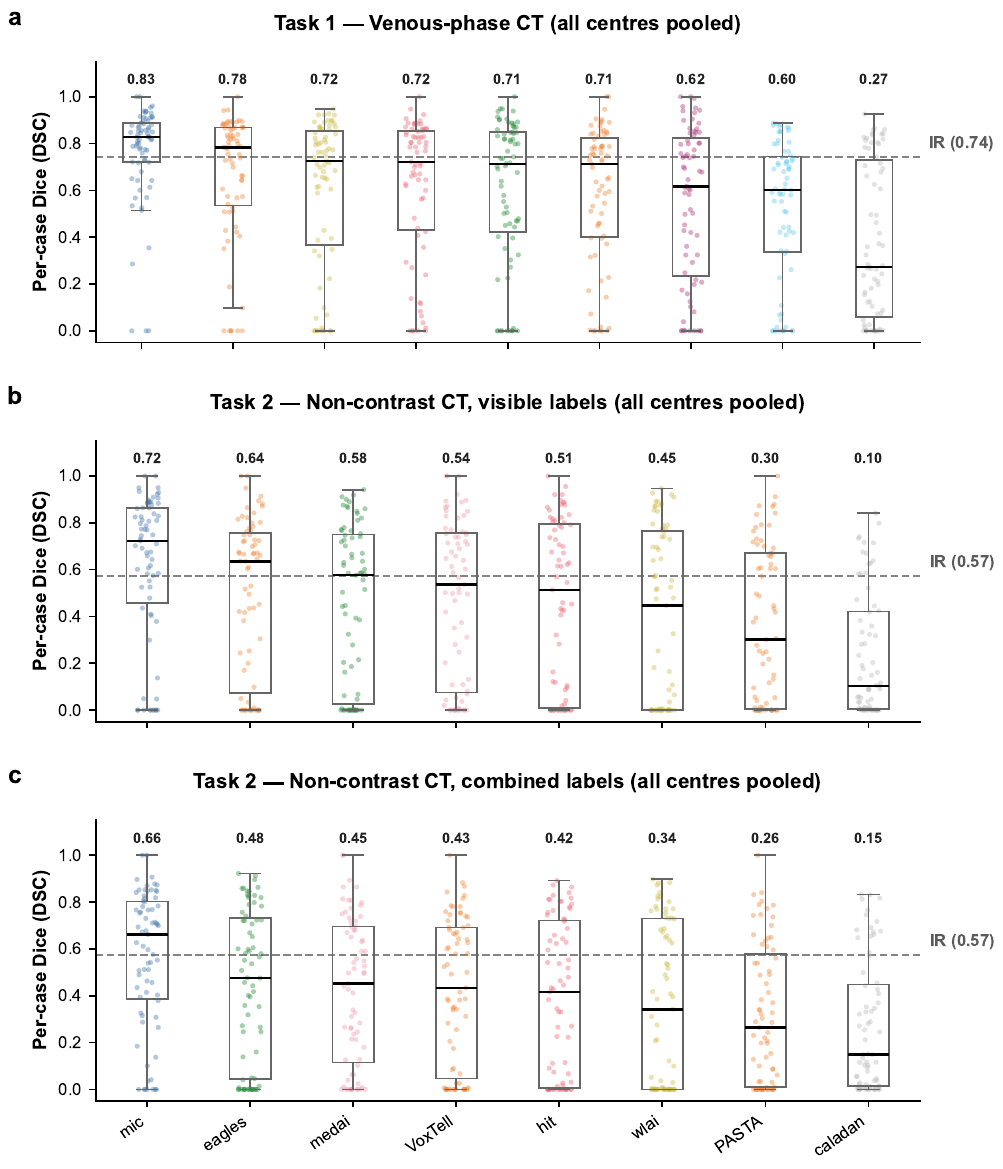}
  \caption{\textbf{Per-case Dice distributions for all participating teams and off-the-shelf baselines.}
    Each panel shows per-case Dice pooled across all three test centres ($n = 70$ volumes).
    Transparent boxes span the interquartile range with a thick black line at the median; individual cases are overlaid as jittered points.
    Challenge participants are shown alongside off-the-shelf baselines VoxTell~1.1 and PASTA (orange); dashed lines indicate inter-reader ceilings.
    \textbf{a,} Task~1 (venous CT): \texttt{mic} achieves median 0.83.
    \textbf{b,} Task~2 (NCCT, visible labels): \texttt{mic} median 0.72.
    \textbf{c,} Task~2 (NCCT, combined labels): \texttt{mic} median 0.66. Teams are sorted by median Dice within each panel.}
  \label{fig:violin}
\end{figure}

\begin{figure}[htbp]
  \centering
  \includegraphics[width=\linewidth]{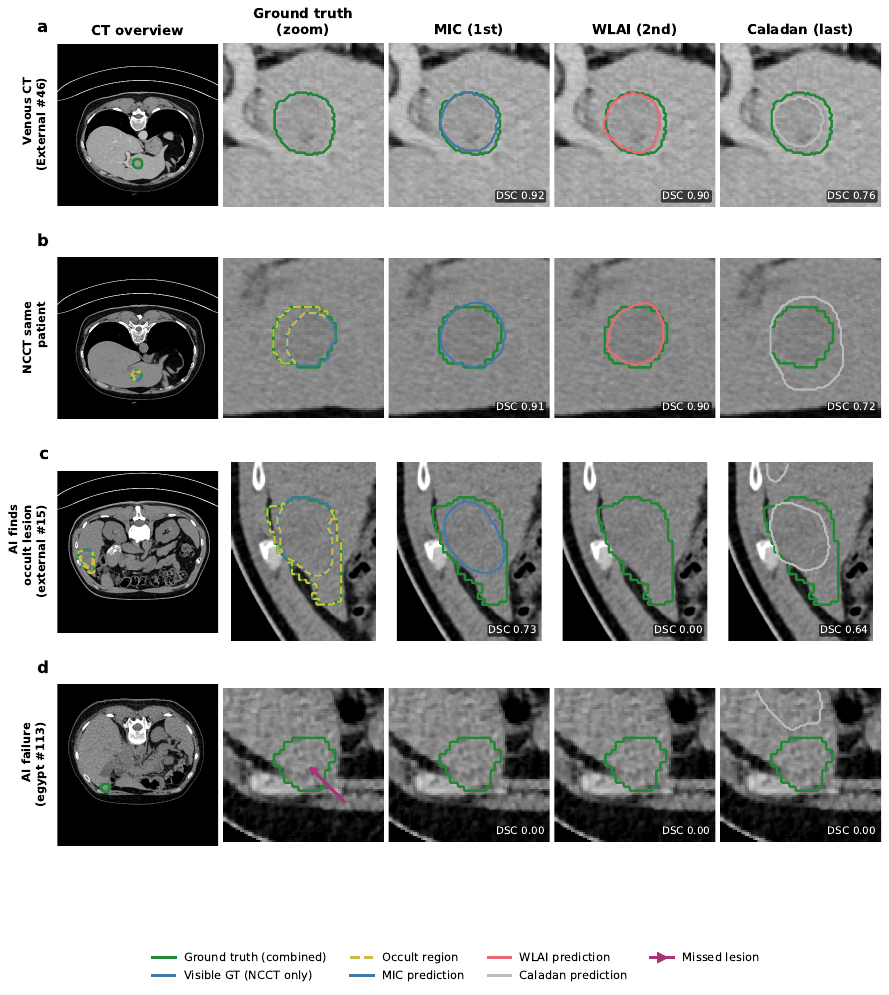}
  \caption{\textbf{Qualitative segmentation comparison across teams (MIC 1st, WLAI 2nd, Caladan last).}
    \textbf{a,} Venous CT (External \#46): all teams segment the lesion accurately (DSC 0.76--0.92).
    \textbf{b,} Same patient, NCCT: the combined GT (green) includes an occult region (yellow dashed) absent from the visible GT (blue). MIC and WLAI maintain DSC~$\geq$~0.90, while Caladan drops to 0.72.
    \textbf{c,} AI detects occult lesion (External \#15): MIC segments an occult lesion with DSC 0.73; WLAI completely fails (DSC 0.00), while Caladan achieves partial detection (0.64). This case has 4 combined-paradigm lesions but only 1 visible on NCCT, demonstrating how multi-phase supervision enables detection of contrast-dependent lesions.
    \textbf{d,} AI failure (Egypt \#113): a small lesion missed by all teams (DSC 0.00), illustrating the sub-100\,mm$^3$ detection gap.}
  \label{fig:qualitative}
\end{figure}

\textbf{Task~1: Venous-phase CT.}
The top-performing method (MIC) achieved per-case Dice scores of 0.721 (China), 0.791 (Egypt), and 0.750 (external), ranking first across all three centres without centre-specific adaptation~(Fig.~\ref{fig:violin}a; Table~\ref{tab:trials2025_t1}). 
Averaging across centres, MIC achieved a mean per-case Dice of 0.754 (median 0.83 pooled across all 70 test volumes). Performance spread was wide (Dice 0.280 to 0.791 on Egypt), with the top three methods clustering within 7\% while lower-ranked teams exhibited high inter-case variance~(Fig.~\ref{fig:violin}a; Fig.~\ref{fig:qualitative}a). Global Dice consistently exceeded per-case Dice~(Table~\ref{tab:trials2025_t1}), indicating that large lesions dominate volumetric overlap and mask failures on small targets.
 
\textbf{Task~2: Non-contrast CT.}
Non-contrast CT was fundamentally harder across all centres and annotation paradigms~(Fig.~\ref{fig:violin}b,c; Table~\ref{tab:trials2025_t2}). Under the combined paradigm, the best per-case Dice reached only 0.586 (China), 0.471 (Egypt), and 0.651 (external), a 10\% to 32\% absolute reduction relative to venous-phase scores. The visible paradigm, evaluating only against lesions annotated on NCCT alone, yielded higher apparent Dice: 0.621 (China), 0.552 (Egypt), and 0.674 (external). Egypt was consistently the hardest centre, with mean combined Dice across teams of only 0.295, compared with 0.449 in China and 0.511 in the external cohort. No team achieved precision and recall both exceeding 0.8 for Task~2 across all centres, confirming that non-contrast liver lesion detection remains an open problem.
 
Stratified analysis by lesion volume revealed consistent size-dependent patterns~(Supplementary Fig.~\ref{edfig:lesion_size}). For Task~1, lesions smaller than 100\,mm$^3$ achieved a mean per-lesion F1 below 0.35. For Task~2, sub-100\,mm$^3$ lesions were almost never detected (F1~$< 0.15$), underscoring the combined effects of partial-volume averaging and poor lesion-to-parenchyma contrast.
 
\begin{figure}[htbp]
  \centering
  \begin{subfigure}[b]{\linewidth}
    \centering
    \includegraphics[width=\linewidth]{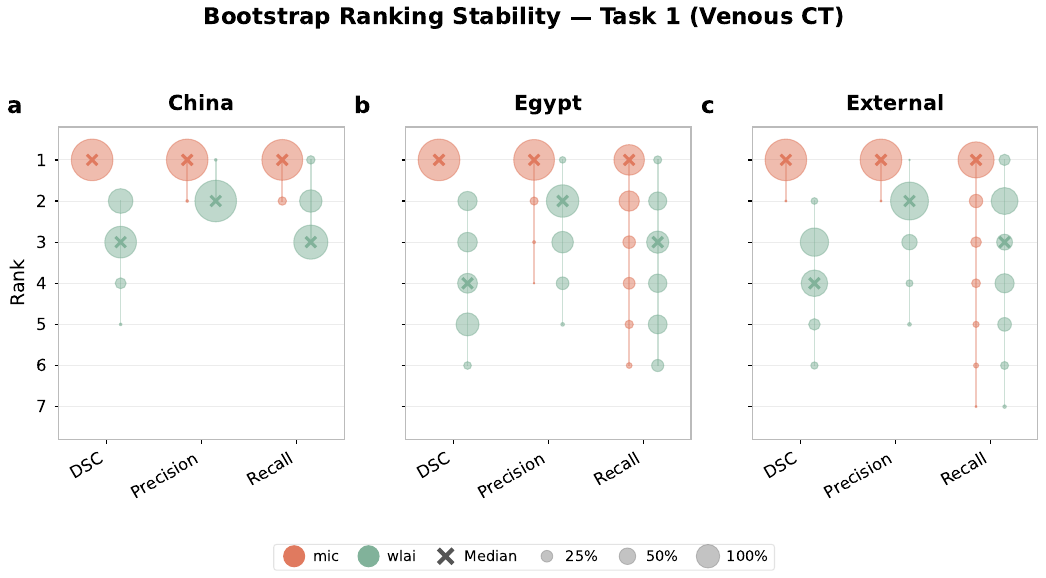}
    \caption{Task~1 (venous CT). Per-case scores were resampled with replacement ($N{=}1{,}000$ iterations) and team rankings recomputed at each iteration.
    Bubble area is proportional to the percentage of iterations achieving a given rank; $\times$ marks the median rank.
    \textbf{Left,}~China. \textbf{Centre,}~Egypt. \textbf{Right,}~External.
    \texttt{mic} achieves rank~1 in 100\% of bootstrap iterations across all centres and metrics.}
    \label{edfig:bootstrap_t1}
  \end{subfigure}
  \vspace{0.5em}
  \begin{subfigure}[b]{\linewidth}
    \centering
    \includegraphics[width=\linewidth]{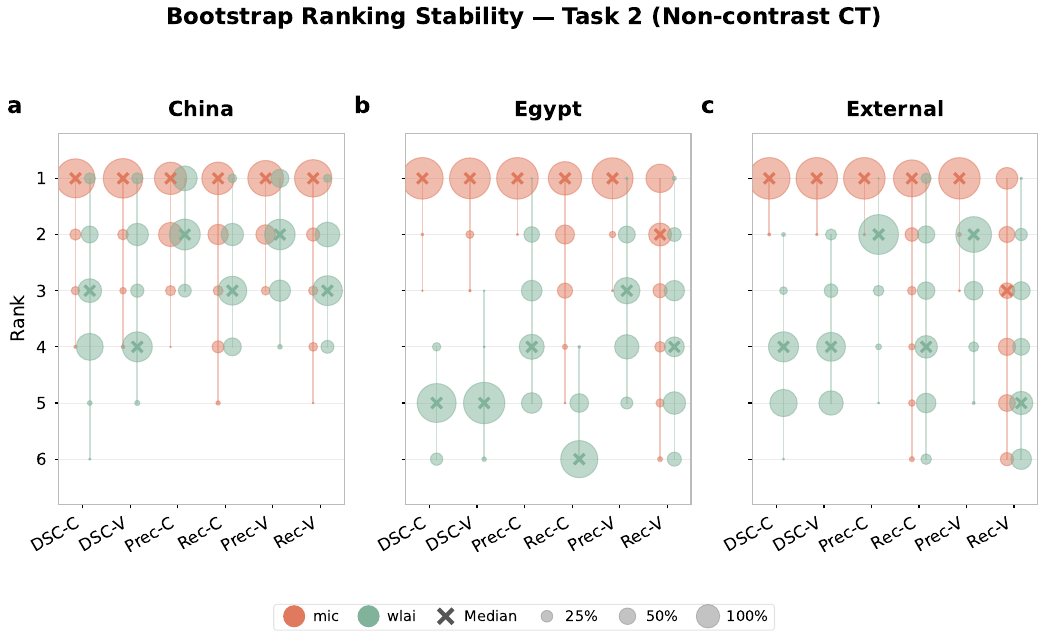}
    \caption{Task~2 (non-contrast CT). Same resampling protocol, applied across six participating teams.
    Combined (C) and visible (V) annotation paradigms are evaluated separately.
    \textbf{Left,}~China. \textbf{Centre,}~Egypt. \textbf{Right,}~External.
    \texttt{mic} maintains rank~1 in the majority of iterations; \texttt{wlai} shows greater variability, particularly on the Egyptian cohort.}
    \label{edfig:bootstrap_t2}
  \end{subfigure}
  \caption{\textbf{Bootstrap ranking stability.}
    Rankings are computed across all participating teams; only the top two (\texttt{mic} and \texttt{wlai}) are shown.}
  \label{fig:bootstrap}
\end{figure}

\textbf{Ranking stability.}
Bootstrap resampling of per-case test scores ($N = 1{,}000$ iterations) confirmed that top-4 rankings were stable, with 95\% confidence interval overlap below 10\%~(Fig.~\ref{fig:bootstrap}). MIC consistently ranked first across all bootstrap iterations for Task~1 on all three centres and maintained rank~1 in the majority of iterations for Task~2~( Figs.~\ref{edfig:bootstrap_t1},~\ref{edfig:bootstrap_t2}).

\subsection*{Comparison with off-the-shelf models and drivers of performance}
 
\begin{figure}[htbp]
  \centering
  \includegraphics[width=\linewidth]{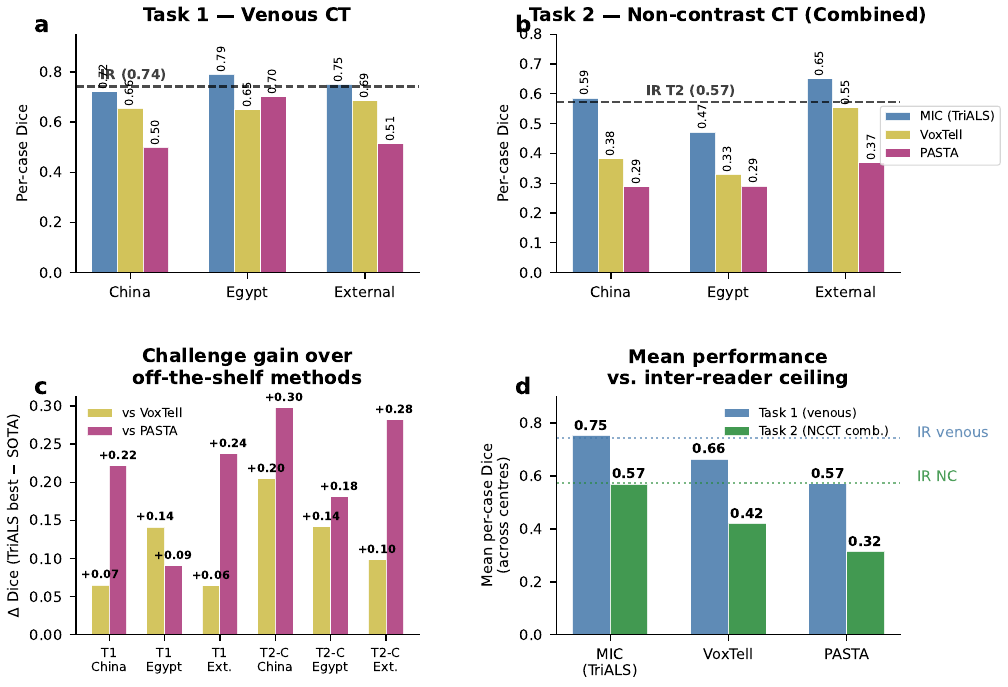}
  \caption{\textbf{Comparison of TriALS 2025 winner with state-of-the-art off-the-shelf methods.}
    \textbf{a,} Task~1 (venous CT) per-case Dice per centre: \texttt{mic\_v2} (TriALS), VoxTell~1.1, and PASTA, with venous inter-reader ceiling (dashed line at 0.742).
    \textbf{b,} Task~2 combined (non-contrast) per-case Dice with NCCT inter-reader ceiling (dashed line at 0.574).
    \textbf{c,} Absolute Dice gain of \texttt{mic\_v2} over each off-the-shelf method per configuration.
    \textbf{d,} Mean per-case Dice across centres overlaid with inter-reader ceilings; \texttt{mic\_v2} operates near the venous ceiling (0.754 vs.\ 0.742).}
  \label{fig:sota}
\end{figure}

\begin{figure}[htbp]
  \centering
  \includegraphics[width=\linewidth]{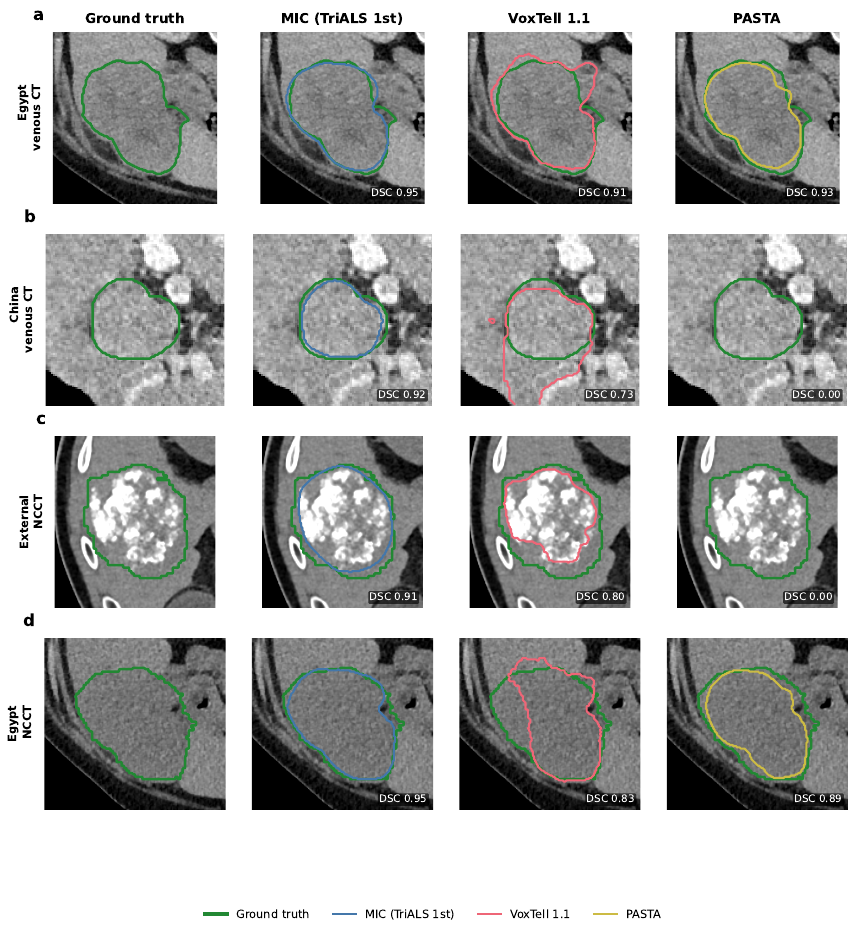}
  \caption{\textbf{Qualitative comparison: TriALS winner vs.\ off-the-shelf methods.}
    \textbf{a,}~Egypt venous (case~101): all methods segment the large lesion, but VoxTell under-segments (DSC~0.91 vs.\ MIC~0.95).
    \textbf{b,}~China venous (case~14): MIC achieves DSC~0.92; VoxTell produces false positives (0.73) and PASTA misses the lesion entirely (0.00).
    \textbf{c,}~External NCCT (case~33): MIC maintains DSC~0.91 on non-contrast; VoxTell under-segments (0.80) and PASTA fails completely (0.00).
    \textbf{d,}~Egypt NCCT (case~101, same patient as~\textbf{a}): MIC achieves DSC~0.95 on NCCT, outperforming both VoxTell (0.83) and PASTA (0.89), demonstrating robust cross-modality generalisation.}
  \label{fig:sota_qual}
\end{figure}

Two contemporary liver lesion segmentation models, VoxTell~\cite{rokuss2025voxtell} and PASTA~\cite{lei2026syntheticdatadrivenradiologyfoundation_pastas}, were evaluated on all TriALS test sets using publicly released weights~(Fig.~\ref{fig:sota}). Notably, VoxTell's released weights include TriALS training data, giving it a data advantage over challenge participants, who trained on the provided 80 cases alongside varying amounts of external data; PASTA was evaluated without any TriALS-specific fine-tuning. On venous CT, the top TriALS method outperformed VoxTell by 14.1\% and 6.4\% in absolute Dice on Egypt and the external set, respectively, and PASTA by 9.1\% and 23.7\% on the same cohorts~(Fig.~\ref{fig:sota}a; Fig.~\ref{fig:sota_qual}). PASTA exhibited a characteristic generalisation failure: it scored 0.700 on the Egyptian cohort but dropped to 0.513 on the external set, suggesting overlap with its training distribution.

On non-contrast CT, the performance gap was even more pronounced~(Fig.~\ref{fig:sota}b). The best TriALS method exceeded PASTA by 28.2\% and VoxTell by 9.8\% in absolute Dice on the independent external test set; on Egypt, the margins were 18.1\% and 14.1\%, respectively. Despite PASTA being designed for non-contrast CT, its combined Dice remained below 0.37 across all centres. Multi-phase training strategies developed within TriALS provided clinically meaningful gains over both approaches~(Fig.~\ref{fig:sota}c,d).

\begin{figure}[htbp]
  \centering
  \includegraphics[width=\linewidth]{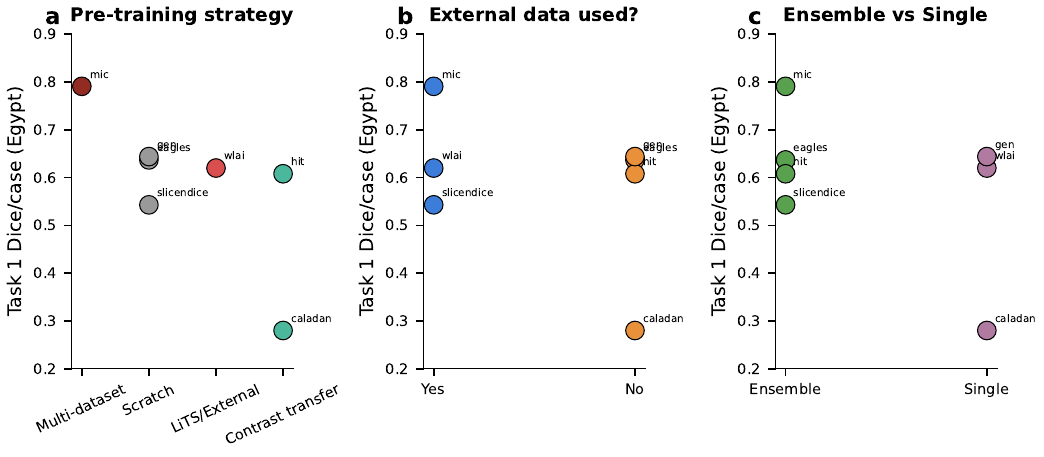}
  \caption{\textbf{Relationship between methodological choices and performance (TriALS 2025, Egypt Task~1).}
    \textbf{a,}~Per-case Dice stratified by pre-training strategy. The multi-dataset pre-training approach (MIC) achieved the highest performance.
    \textbf{b,}~Effect of external data usage: teams that incorporated external datasets tended to achieve higher Dice, though with notable exceptions.
    \textbf{c,}~Ensemble versus single-model inference: ensemble approaches showed no systematic advantage, suggesting that architecture and training strategy were more influential.}
  \label{fig:supp_strategy}
\end{figure}

Beyond model-level comparisons, we examined which methodological choices most influenced ranking~(Fig.~\ref{fig:supp_strategy}). The single strongest predictor was multi-dataset pre-training~(Fig.~\ref{fig:supp_strategy}a). Teams that incorporated external datasets tended to achieve higher Dice, though with notable exceptions~(Fig.~\ref{fig:supp_strategy}b). Ensemble approaches showed no systematic advantage over single-model inference, suggesting that training strategy was more influential than inference-time aggregation~(Fig.~\ref{fig:supp_strategy}c).

\subsection*{The modality gap and cross-year progress}

\begin{figure}[htbp]
  \centering
  \includegraphics[width=\linewidth]{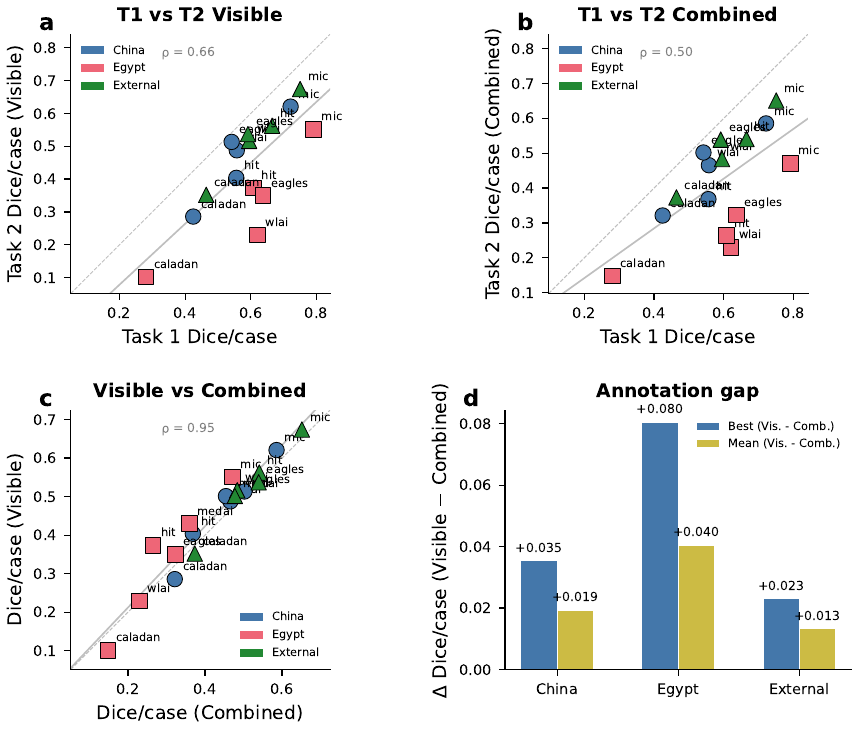}\\[0.5em]
  \includegraphics[width=\linewidth]{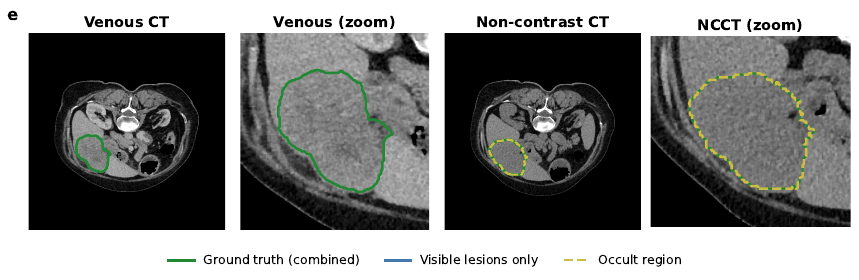}
  \caption{\textbf{The modality gap: cross-task degradation and annotation paradigm comparison.}
    \textbf{a,} Task~1 versus Task~2 visible Dice scatter; all points fall below the diagonal, confirming non-contrast CT is harder regardless of venous-phase capability.
    \textbf{b,} Task~1 versus Task~2 combined Dice scatter.
    \textbf{c,} Combined versus visible Dice scatter per centre.
    \textbf{d,} Summary gap ($\Delta = $ visible $-$ combined Dice) per centre; Egypt shows the largest gap ($\Delta = 0.080$), reflecting the highest occult lesion burden.
    \textbf{e,} Side-by-side comparison of the same patient (Egypt \#101) in venous-phase (left) and non-contrast CT (right). The lesion is clearly delineated on venous CT (green contour) but nearly isodense with surrounding parenchyma on NCCT, where the entire lesion region is classified as occult (yellow dashed = combined GT absent from visible GT), illustrating the fundamental contrast-dependent visibility that drives the modality gap.}
  \label{fig:degradation}
\end{figure}

The difference between visible-paradigm and combined-paradigm Dice ($\Delta = 0.023$--$0.080$) quantifies the burden of occult lesions per centre~(Fig.~\ref{fig:degradation}c--e). Egypt showed the largest gap ($\Delta = 0.080$), consistent with its higher proportion of small, contrast-dependent lesions. A positive correlation between Task~1 and Task~2 visible Dice was observed across teams and centres (Spearman $\rho = 0.66$, 95\% bootstrap CI [0.11, 0.92]; $p = 0.008$; $n = 15$ team--centre pairs), suggesting that venous-phase capability partially predicts non-contrast performance~(Fig.~\ref{fig:degradation}a--b). However, the regression slope was less than unity, confirming that non-contrast CT introduces additional challenge beyond contrast-enhanced segmentation difficulty. The clinical implication is direct: AI deployed in contrast-limited settings should not be benchmarked solely on contrast-enhanced data.
 
\begin{figure}[htbp]
  \centering
  \includegraphics[width=\linewidth]{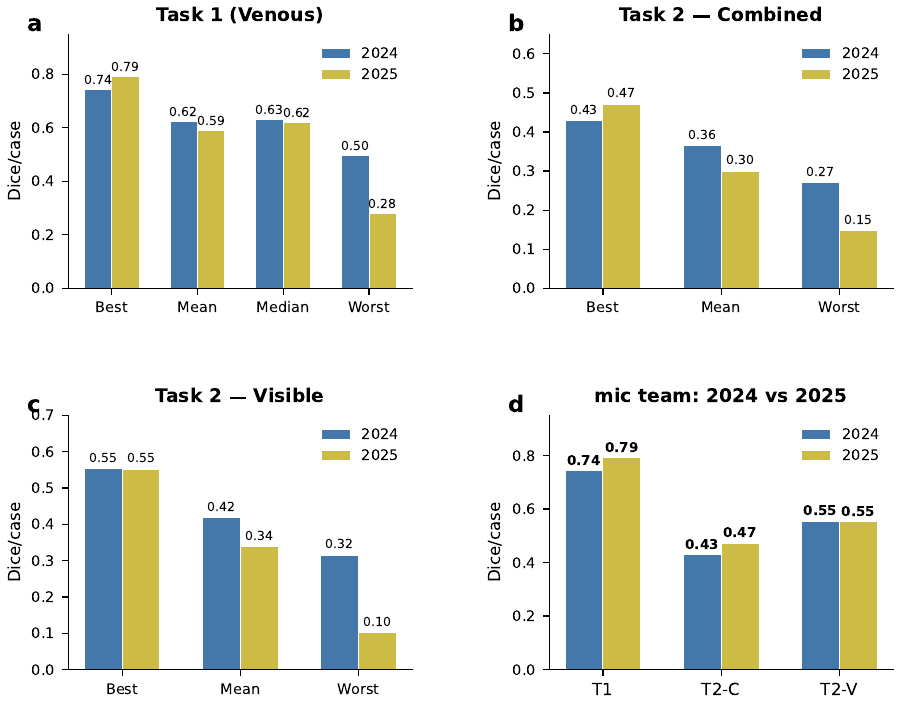}
  \caption{\textbf{Cross-year progress (Egypt test set; 2024 vs.\ 2025).}
    \textbf{a,} Task~1: best Dice improved from 0.742 to 0.791 (+4.9\%).
     \textbf{b,} Task~2 combined Dice improved from 0.429 to 0.471 (+4.2\%).
     \textbf{c,} Task~2 visible Dice essentially unchanged (0.55 → 0.55).
    \textbf{d,} \texttt{mic} team trajectory across all six settings, showing gains concentrated in the combined paradigm.
    }
  \label{fig:cross_year}
\end{figure}

Comparing results on the shared Egyptian test set across challenge editions reveals measurable improvement at the top end~(Fig.~\ref{fig:cross_year}; Table~\ref{tab:trials2024}). The best Task~1 per-case Dice improved from 0.742 (2024) to 0.791 (2025), a 4.9\% absolute gain driven by the expansion of MultiTalent~\cite{ulrich2023multitalent} pre-training from 30 to 86 public datasets. Task~2 revealed an asymmetric pattern: combined Dice improved from 0.429 to 0.471, whereas visible Dice remained essentially unchanged (0.553 versus 0.552). The combined paradigm benefits from multi-phase training strategies that propagate contrast-derived labels to non-contrast inference. However, the visible paradigm depends on intrinsic non-contrast features, and the persistent plateau suggests a fundamental perceptual barrier that scaling pre-training data alone cannot overcome, pointing to the need for qualitatively different approaches to non-contrast lesion detection.

\section*{Discussion}


A central finding of this work is that non-contrast CT is not merely a lower-quality variant of venous-phase imaging but a fundamentally harder modality requiring distinct computational strategies. This distinction is clearly reflected in TriALS results: the top-performing method achieved a mean venous-phase Dice of 0.754 across three test centres, consistent with human-level performance, yet dropped to a mean of 0.57 on NCCT. On NCCT, methods that trained jointly on both contrast and non-contrast phases achieved absolute Dice gains of up to 28.2\% over PASTA~\cite{lei2026syntheticdatadrivenradiologyfoundation_pastas} and 9.8\% over VoxTell~\cite{rokuss2025voxtell} on the independent external test set. Multi-phase supervision effectively transfers radiological knowledge from contrast images to guide non-contrast inference. Despite these gains, no team solved small-lesion detection on non-contrast CT: below 100\,mm$^3$, per-lesion F1 remained under 0.15 across all submissions. Partial-volume effects and low signal-to-noise ratios render such lesions nearly indistinguishable from surrounding parenchyma. This size dependence is clinically important because early detection of small metastases directly alters staging and management decisions.


A particularly revealing finding emerges from the cross-year comparison. Between the 2024 and 2025 editions, the top method's combined-paradigm Dice improved from 0.429 to 0.471, yet its visible-paradigm Dice remained essentially unchanged (0.553 to 0.552). This asymmetry distinguishes two fundamentally different mechanisms: multi-phase supervision enables \textit{knowledge transfer}, allowing algorithms to predict lesion locations learned from contrast-phase training data even when those lesions are invisible on the input non-contrast image. However, it does not enhance \textit{perceptual ability}, the capacity to detect lesions from intrinsic NCCT features alone. Because the winning team (MIC) was the only team to participate in both editions, this improvement reflects one team's methodological refinement, specifically the expansion of pre-training from 30 to 86 public datasets, rather than a community-wide trend. Nevertheless, the persistent visible-paradigm plateau suggests that current architectures have reached a perceptual barrier on non-contrast CT that cannot be overcome by scaling pre-training data alone, and that qualitatively different approaches, such as texture-aware encoders, diffusion-based contrast synthesis, or interactive human-in-the-loop refinement, may be required to advance beyond this ceiling.


These results carry direct implications for clinical deployment in resource-constrained environments. Artificial intelligence tools intended for contrast-limited settings cannot be validated on contrast-enhanced data alone; doing so would overestimate real-world performance and mask failure modes specific to non-contrast acquisition. Our combined annotation paradigm, in which venous-phase delineations guide non-contrast labels, demonstrates a practical route for transferring expert knowledge across imaging phases. This paradigm is epistemologically distinct from standard segmentation: algorithms must predict lesion locations that are not directly visible on the input image, guided by patterns learned during multi-phase training. This is closer to a lesion prediction task than boundary delineation, which is why we evaluate the combined paradigm primarily with detection-oriented metrics rather than surface distance measures (Methods). Even imperfect automated assistance may improve on the 0.333 lesion-level precision that expert radiologists achieved when annotating non-contrast CT in our inter-reader study. In many low-resource hospitals across Africa and Asia, radiologists lack access to contrast agents entirely owing to cost or supply-chain constraints. For those settings, a tool that surfaces candidate lesions for rapid radiologist review could meaningfully reduce missed diagnoses. Interactive segmentation offers a complementary path forward~\cite{interactive}: a radiologist providing a few clicks or scribbles can refine algorithmic proposals within seconds, and this human-in-the-loop paradigm may prove more realistic and safer than fully automatic pipelines for non-contrast CT.


From a methodological standpoint, the near-universal adoption of nnU-Net~\cite{isensee2021nnunet} across participating teams confirms the framework's versatility but means that the solution space explored by this challenge was narrower than ideal. Crucially, this architectural convergence did not limit clinical value: the top submission outperformed two contemporary off-the-shelf models by 6--24\% absolute Dice on venous CT and by up to 28\% on NCCT~(Fig.~\ref{fig:sota}; Fig.~\ref{fig:sota_qual}), demonstrating that training strategy, data curation, and multi-phase supervision yield gains well beyond what architecture choice alone provides. Foundation models for medical image segmentation~\cite{SAM,Ma2024} were not explored by any team despite their potential for zero-shot generalisation, and centre-aware adaptation through domain tokens or site-specific normalisation layers remains an unexploited opportunity. Future editions of TriALS could incentivise architectural diversity through dedicated tracks or by providing pre-computed features that lower the barrier for non-nnU-Net approaches.

A further methodological consideration is registration error in the combined labels. Aligning venous and non-contrast volumes introduces spatial uncertainty that propagates into ground-truth boundaries. Several design choices mitigate this risk: registration was performed on liver-cropped volumes rather than full abdominal scans; sequential rigid-to-non-rigid registration via elastix~\cite{elastix} resolved residual deformations from respiratory motion; STAPLE-based label fusion~\cite{Warfield2004} across all four phases filtered out poorly registered single-phase artefacts; and an abdominal radiology consultant verified each combined-only lesion against the original contrast-enhanced images, correcting or excluding cases with visible registration errors. Surface distance metrics were excluded from combined-paradigm ranking to ensure that residual registration noise does not bias the primary clinical metric. All raw multi-phase volumes and per-phase annotations are publicly released to enable independent verification.


Several limitations temper these conclusions. First, the dataset spans only two countries (China and Egypt) plus one external site; representation from Latin America, sub-Saharan Africa, and South Asia, where contrast shortages are arguably most severe, is absent. We view TriALS as a necessary first step; broader initiatives such as African and pan-Asian collaborations will be needed to validate these findings across the full diversity of populations and clinical workflows. Second, with 150 cases and test sets of $n{=}17$--28, the benchmark is modest by modern standards. Bootstrap confidence interval widths of 0.13--0.27 for the top method's per-centre Dice are sufficient to detect the large gaps between tasks and between top- and bottom-ranked teams, but potentially underpowered for fine-grained cross-centre comparisons; non-significant cross-centre $p$-values should be interpreted as inconclusive rather than evidence of equivalence. Third, the inter-reader study comprised only 16 cases per phase evaluated by two radiologists, yielding a human ceiling estimate with a wide confidence interval. While consistent with prior challenge substudies~\cite{Lits_benchmark,MAIERHEIN2020101796}, the bimodal agreement pattern means that the ceiling should be interpreted as an approximate reference rather than a precise threshold. Fourth, the dataset lacks lesion-type annotations (e.g., hepatocellular carcinoma vs.\ metastases vs.\ benign entities); stratified analysis by tumour type would be informative for clinical deployment~\cite{Wu2025}. Finally, reducing inference latency through model distillation or lighter architectures is an important challenge that the current benchmark does not address.


Looking ahead, several extensions follow naturally. Triphasic joint exploitation, combining arterial, venous, and delayed phases, could further boost performance and enable lesion characterisation. Longitudinal datasets would support treatment-response assessment. Expanding to additional regions and scanner vendors would strengthen generalisability claims. A dedicated interactive-segmentation track would directly evaluate human-in-the-loop workflows that may represent the most practical path to clinical adoption. Finally, the perceptual barrier identified here defines a clear research frontier: overcoming the visible-paradigm plateau on non-contrast CT will require approaches that go beyond scaling existing architectures and training data, potentially drawing on contrast synthesis, self-supervised texture learning, or foundation models adapted for low-contrast imaging---advances that would directly benefit the contrast-limited healthcare systems TriALS was designed to serve.
\section*{Methods}

\subsection*{Study design and ethical approval}

TriALS was organised as a retrospective, multi-centre biomedical image analysis challenge following the BIAS (Biomedical Image Analysis ChallengeS) reporting guidelines~\cite{MAIERHEIN2020101796} and metric-selection recommendations from Reinke et al.~\cite{Reinke2024}. Imaging data were collected from routine clinical examinations at two primary participating institutions. The Egyptian dataset (352 volumes from Ain Shams University Hospitals, Cairo, Egypt) was collected under ethical approval from the local Research Ethics Committee (FWA~000017585), with waiver of individual informed consent approved for retrospective de-identified data use. The Chinese dataset (148 volumes from Sun Yat-Sen Memorial Hospital, Sun Yat-Sen University, Guangzhou, China) was collected under approval from the institutional review committee of Sun Yat-sen Memorial Hospital, Sun Yat-sen University. All imaging data were fully anonymized prior to transfer; direct patient identifiers, dates, and institution-identifying metadata were removed or shifted. No additional data collection beyond routine clinical care was required.

The external test set was sourced from Nanfang Hospital, Southern Medical University (Guangzhou, China), an institution not represented in the training data, to provide an independent assessment of generalization. Ethical approval for this external dataset was obtained locally at the contributing institution prior to data transfer to the challenge organizers.

\subsection*{Study population and inclusion criteria}

The challenge cohort included adult patients ($\geq18$ years) who underwent complete four-phase abdominal CT (non-contrast, arterial, portal venous, delayed) at either participating institution. Inclusion required: (1) at least one liver lesion reported in the clinical radiology report; (2) all four imaging phases available; (3) acceptable image quality on visual assessment. Exclusion criteria were: artifacts significantly affecting diagnostic quality; total main portal vein thrombosis; and advanced cirrhosis grade C (Child-Pugh) on CT assessment. The resulting dataset reflects a representative spectrum of hepatic lesion types encountered in abdominal radiology practice, including hepatocellular carcinoma, liver metastases, intrahepatic cholangiocarcinoma, and benign entities.

\subsection*{CT acquisition protocols}

CT acquisitions followed standard institutional abdominal oncology protocols. The imaging was performed on eight CT scanner models spanning three manufacturers: SIEMENS SOMATOM Force, SIEMENS SOMATOM Sensation~64, United Imaging uCT780, GE Discovery CT750~HD, GE Revolution~EVO, GE Healthcare Optima, Toshiba Prime Aquilion, and GE BrightSpeed~CT. Contrast-enhanced phases were acquired after intravenous injection of iodinated contrast agent: arterial phase upon bolus tracking in the descending aorta (approximately 35–45~s post-injection), portal venous phase approximately 20~s after arterial phase (65–70~s post-injection), and delayed phase at 3~minutes post portal venous phase. In the Egyptian cohort, contrast agents were variable due to supply-chain limitations; non-ionic contrast (Omnipaque) was used when available, with ionic contrast (Telebrix) used when necessary.

Acquisition parameters varied by institution and scanner model but spanned: tube voltage 100–120~kVp; tube current modulated to local dose reference levels; slice thickness 0.63–1.56~mm (training mean $0.97\pm0.19$~mm); in-plane pixel spacing 0.54–1.19~mm (training mean $0.832\pm0.10$~mm); matrix $512\times512$ throughout; field of view covering the full abdomen. All volumes were reconstructed using soft-tissue or mixed reconstruction kernels; no edge-enhancing kernels were included. For multi-phase acquisitions, reconstruction was performed at the same slice thickness across all phases when possible. Detailed per-cohort acquisition statistics are summarized in Table~\ref{tab:cohort}.

\subsection*{Scanner and protocol effects.}
Volumes acquired on Siemens SOMATOM platforms (predominantly Egypt) exhibited higher noise floors in the non-contrast phase relative to United Imaging and GE scanners used in China and the external site~(Table~\ref{tab:cohort}; Supplementary Fig.~\ref{fig:spacing}). In-plane resolution differences (Egypt mean 0.862$\pm$0.095\,mm versus China mean 0.766$\pm$0.063\,mm versus External mean 0.686$\pm$0.072\,mm) contributed to partial-volume averaging that particularly affected small-lesion segmentation. Volume depth varied by more than two-fold (Egypt 578$\pm$137 versus External 240$\pm$69 slices), reflecting different scan coverage protocols. These acquisition differences were statistically significant across all centre pairs (Mann--Whitney $U$, $p < 0.006$ for 8 of 9 comparisons; Table~\ref{tab:centre_pvalues}). Yet despite these measurable effects, segmentation performance differences across centres were not statistically significant for Task~1 Dice ($p > 0.20$), suggesting that current methods are reasonably robust to acquisition heterogeneity on venous-phase data. We note that the segmentation-metric centre comparisons in Table~\ref{tab:centre_pvalues} use team-level scores ($n = 6$--$7$ teams) rather than per-case scores, limiting statistical power; the non-significant results should be interpreted with this caveat. No participating method explicitly modelled scanner vendor or centre identity, suggesting that centre-aware adaptation remains an unexploited opportunity that could yield additional gains on non-contrast data.

\subsection*{Annotation procedure}

Lesion annotations were produced following a two-stage hybrid human–algorithm protocol. In the first stage, an automated algorithm trained on LiTS~\cite{Lits_benchmark} and a small set of in-house manually annotated pilot cases generated preliminary segmentations for all volumes. Non-rigid registration was applied to align the contrast-enhanced phases to the non-contrast CT, and lesion masks were propagated from the phase of highest lesion conspicuity to all other phases. These automatic predictions were not used directly as ground truth; they served solely to accelerate the manual annotation workflow.

In the second stage, a team of 10 trained annotators (radiology residents and fellows with a minimum of one year of focused abdominal imaging experience) reviewed and corrected all automatic segmentations using 3D Slicer. Annotators received standardised training via recorded video demonstrations and participated in troubleshooting sessions with abdominal radiology consultants. Lesions spanning at least two consecutive CT slices (approximately $\geq3$~mm in maximum diameter given a typical slice thickness of $1.25$~mm) were included. Each annotation was subsequently reviewed by a second annotator; disagreements were escalated to a senior radiologist consultant (${>}10$ years of experience) for consensus resolution. Ambiguous boundary cases and lesions only visible in a single phase were flagged for additional review.

\textbf{Inter-rater variability.} To quantify annotation uncertainty, two abdominal radiologists independently annotated a randomly selected subset of cases from the combined Egyptian and Chinese cohorts ($n{=}16$ venous; $n{=}16$ non-contrast). Inter-rater per-case Dice for venous-phase annotations was $0.742\pm0.443$ (mean $\pm$ SD), with lesion-level detection precision of 0.833 and recall of 0.946, consistent with published inter-observer agreement for liver lesion segmentation. For non-contrast CT, per-case Dice was substantially lower ($0.574\pm0.490$, precision 0.333, recall 0.769), confirming the fundamental difficulty of NCCT annotation. The high standard deviations reflect a bimodal agreement pattern: radiologists either agree perfectly on lesion delineation (DSC\,$\approx$\,1) or disagree completely on lesion visibility (DSC\,$=$\,0). These per-case Dice values establish principled empirical performance ceilings for automated methods under each imaging paradigm.

\subsection*{Label fusion and combined annotation paradigm}

For training, each CT case contains five annotation sets: individual lesion segmentations for each of the four phases, plus a combined annotation registered to the non-contrast phase. The combined annotation was constructed as follows:
\begin{enumerate}[leftmargin=1.5em,itemsep=0pt]
    \item \textbf{Rigid registration:} Annotations from arterial, portal venous, and delayed phases were aligned to the non-contrast CT coordinate frame using rigid registration via elastix~\cite{elastix} to correct for respiratory motion and patient repositioning between phases.
    \item \textbf{Non-rigid refinement:} Phase-specific deformations were addressed using non-rigid registration via elastix, proceeding sequentially from rigid to non-rigid transforms to achieve sub-voxel alignment of lesion boundaries.
    \item \textbf{Label fusion:} The combined label was generated using the Simultaneous Truth and Performance Level Estimation (STAPLE) algorithm~\cite{Warfield2004} via SimpleITK, fusing all four single-phase annotations to produce a probabilistic consensus, ensuring that any lesion visible in at least two phases was included in the combined mask.
    \item \textbf{Quality assurance:} The resulting combined annotations were reviewed by an abdominal radiology consultant to verify that the fused mask accurately represented the complete lesion burden across modalities.
\end{enumerate}

For test and validation sets, two evaluation paradigms were defined. The \textit{visible paradigm} evaluates only against the non-contrast single-phase annotation, assessing the ability to detect lesions visible to the radiologist without contrast information. The \textit{combined paradigm} evaluates against the multi-phase fused annotation, assessing the ability to detect the complete lesion burden, including lesions invisible in NCCT alone—the primary clinical goal of the challenge.

\subsection*{Data split and pre-processing}

The dataset was divided into training ($n{=}80$: 60~Egypt, 20~China) and three test sets: Egypt internal ($n{=}28$), China internal ($n{=}17$), and external ($n{=}25$). Test set labels were withheld until after the final submission deadline.

All CT volumes were distributed in their native clinical format (NIfTI conversion from DICOM, no resampling or intensity normalization applied by the organizers). Voxel intensities were retained in Hounsfield units. This design choice places the responsibility for pre-processing—including intensity windowing, normalization, and optional resampling—on participating teams, reflecting the challenge of real-world heterogeneous data and encouraging innovation in domain-adaptive pre-processing strategies. Baseline code provided by the organizers included a reference pre-processing pipeline based on nnU-Net~\cite{isensee2021nnunet} defaults (foreground-adapted intensity normalization, anisotropy-aware resampling), which teams were free to adopt or replace.

\subsection*{Challenge organisation and submission procedure}

The challenge was hosted on the Synapse.org platform (\url{https://www.synapse.org/\#!Synapse:syn53285416}), with code available at \url{https://github.com/xmed-lab/TriALS}. The challenge followed a two-phase structure: a development phase (April–August 2025) during which teams downloaded training data and received informal feedback on the validation set; and a final evaluation phase (September 2025) in which teams submitted Docker containers via Synapse. Each container was executed on standardised hardware (NVIDIA RTX~3090 24\,GB GPU, 128~GB RAM, 32 CPU cores) by the organizers. Teams were permitted a maximum of three validation submissions and one final test submission per task. Organizer-affiliated teams participated but were ineligible for awards and excluded from the public leaderboard. To ensure evaluation integrity, the following safeguards were implemented: (1) test set labels were generated and held by a separate data-management team and were not accessible to any participating team, including organizer-affiliated teams, until after the submission deadline; (2) all submissions, including those from organizer-affiliated teams, were evaluated through the same automated Synapse.org pipeline with no manual intervention; (3) organizer-affiliated teams were subject to the same submission limits (three validation, one final test) as all other participants; and (4) final rankings and all analyses reported in this paper were computed after the submission deadline, with no opportunity for post-hoc optimisation.

The evaluation server computed all metrics against the hidden test labels and updated the leaderboard following each submission. Final rankings were computed only after the submission deadline closed, preventing post-hoc test-set optimization. Top-ranked teams were invited to submit detailed method descriptions and participate in the joint analysis comprising this paper.

\subsection*{Evaluation metrics}

Performance was evaluated using a suite of lesion-level segmentation and detection metrics, computed using \texttt{medpy}\footnote{\url{https://github.com/loli/medpy}} for spatial overlap and a custom \texttt{Surface} class for distance-based measures. Predicted and reference volumes were first decomposed into connected components; a lesion-matching algorithm then paired predicted and reference lesions, merging labels where a single prediction spanned multiple reference lesions and vice versa.

\textbf{Segmentation metrics} (computed per matched lesion, then averaged across all lesions including penalties):
\begin{enumerate}[leftmargin=1.5em,itemsep=2pt]
    \item \textbf{Dice Similarity Coefficient (DSC).} Computed per-lesion as $\text{Dice} = 2|P \cap G| / (|P| + |G|)$. Two dataset-level variants are reported: per-case Dice (mean of per-volume binary Dice values) and global Dice ($2 \sum I / \sum S$ across all volumes).
    \item \textbf{Volumetric Overlap Error (VOE).} $\text{VOE} = 1 - \text{Jaccard}$, where $\text{Jaccard} = \text{Dice} / (2 - \text{Dice})$.
    \item \textbf{Relative Volume Difference (RVD).} Absolute relative volume difference between prediction and reference, ranked by absolute value.
    \item \textbf{Average Symmetric Surface Distance (ASSD).} Mean bidirectional surface distance between predicted and reference lesion surfaces.
    \item \textbf{Root Mean Square Symmetric Surface Distance (RMSD).} Root mean square of the bidirectional surface distances.
    \item \textbf{Maximum Symmetric Surface Distance (MSD).} Hausdorff distance between prediction and reference surfaces.
\end{enumerate}

\textbf{Detection metrics} (computed at dataset level from aggregated true-positive, false-positive, and false-negative counts):
\begin{enumerate}[leftmargin=1.5em,itemsep=2pt]
    \item \textbf{Precision and Recall (IoU~$> 0.5$).} A predicted lesion is counted as a true positive if its intersection-over-union (IoU) with the matched reference lesion exceeds 0.5. Precision~$= \text{TP}/(\text{TP}+\text{FP})$; Recall~$= \text{TP}/(\text{TP}+\text{FN})$.
    \item \textbf{Precision and Recall (IoU~$> 0$).} A lenient detection criterion requiring any non-zero voxel overlap between predicted and reference lesions. This quantifies coarse localisation ability independent of segmentation fidelity.
\end{enumerate}

\textbf{Penalty convention.} Undetected reference lesions (false negatives) and spurious predictions (false positives) are assigned worst-case metric values: Dice~$= 0$, Jaccard~$= 0$, VOE~$= 1.0$, and a fixed penalty of $9{,}001$ for RVD, ASSD, RMSD, and MSD. These penalties are included in the per-lesion average, ensuring that skipping lesions is penalised more severely than producing imperfect segmentations. The penalty value of $9{,}001$ exceeds the maximum possible surface distance in any volume in the dataset.

\subsection*{Ranking methodology}

We adopted a rank-then-aggregate strategy following established challenge methodology~\cite{maier-hein_2018_bias}. For each metric, teams were ranked individually (rank~1~=~best); for lower-is-better metrics (VOE, RVD, ASSD, RMSD, MSD), ascending order was used, and for higher-is-better metrics (Dice, precision, recall), descending order was used. RVD was ranked by absolute value. The final leaderboard rank was computed as the mean rank across all included metrics.

The set of metrics used for ranking differed by task and annotation paradigm:

\begin{itemize}[leftmargin=1.5em,itemsep=2pt]
    \item \textbf{Task~1 (venous-phase CT):} Ranked across all 11 metrics (per-case Dice, global Dice, precision and recall at IoU~$> 0.5$, precision and recall at IoU~$> 0$, VOE, RVD, ASSD, RMSD, MSD). This full metric suite captures both lesion detection and segmentation boundary quality.
    \item \textbf{Task~2 --- visible paradigm:} Ranked identically to Task~1 using all 11 metrics, as the visible annotation provides a well-defined spatial reference suitable for surface and volume evaluation.
    \item \textbf{Task~2 --- combined paradigm:} Ranked using only 6 detection-oriented metrics (per-case Dice, global Dice, precision and recall at IoU~$> 0.5$, precision and recall at IoU~$> 0$). Surface distance and volumetric overlap metrics (VOE, RVD, ASSD, RMSD, MSD) were excluded from combined-paradigm ranking. This design reflects the clinical rationale: combined annotations include lesions invisible on non-contrast CT, for which spatial boundary accuracy is less meaningful than detection completeness.
\end{itemize}

For the Task~2 final leaderboard, rankings from the visible and combined paradigms were averaged to produce a single mean rank per team. The rank-then-aggregate approach is metric-scale-invariant and promotes consistent performance across evaluation dimensions rather than optimisation of a single criterion. For missing submissions, worst-case values were assigned (Dice~$= 0$, surface metrics~$= 9{,}001$).

\subsection*{Statistical analysis}

Aggregate statistics are reported as mean $\pm$ standard deviation or median [interquartile range]. Group comparisons (e.g., centre-to-centre) were performed using the Mann–Whitney $U$ test for unpaired comparisons, with a two-sided significance threshold of $\alpha = 0.05$. For the nine pairwise centre comparisons reported in Table~\ref{tab:centre_pvalues}, Benjamini--Hochberg false-discovery-rate (FDR) correction~\cite{fdr} was applied within each parameter family; all acquisition-parameter comparisons reported as significant remain significant after FDR correction at $q = 0.05$. For segmentation metrics, two ASSD comparisons (China vs.\ External, Egypt vs.\ External) that were nominally significant at $p < 0.05$ did not survive FDR correction; these are noted in Table~\ref{tab:centre_pvalues}. Correlation coefficients between tasks are reported as Spearman's $\rho$ with 95\% bootstrap confidence intervals (1{,}000 iterations) to accommodate non-normal distributions. Ranking stability was assessed by bootstrap resampling of the per-case test scores ($N{=}1{,}000$ iterations with replacement), recomputing rankings at each iteration. Percentile-based 95\% confidence intervals are reported for ranking positions and key performance metrics. All analyses were performed in Python~3.10 using NumPy~1.26, SciPy~1.11, scikit-learn~1.3, and Matplotlib~3.8. Statistical significance annotations in figures use: $\ast P < 0.05$; $\ast\ast P < 0.01$; $\ast\ast\ast P < 0.001$; ns $P \geq 0.05$.

\subsection*{Inter-rater agreement measurement}

Inter-rater agreement between the two radiologists was quantified using: (1) per-case Dice coefficient, averaging DSC over all ground-truth lesions in each scan; (2) lesion-level precision and recall using maximum volumetric overlap matching ($\geq 10\%$ IoU threshold); and (3) global Dice computed over all pooled lesion voxels. Lesions annotated by one reader but not the other were counted as false negatives for the second reader and contributed a Dice of~0 to the per-case inter-rater Dice average. This undetected-lesion penalty ensures that the inter-rater Dice values are directly comparable to automated method scores computed under the same convention. Reported inter-rater Dice values—$0.742$ (venous) and $0.574$ (non-contrast)—therefore serve as empirical upper bounds against which challenge methods are benchmarked.


\section*{Data Availability}

The code and model weights for TriALS are publicly available at \url{https://github.com/xmed-lab/TriALS}. The training data and all annotations are hosted on Hugging Face at~\url{https://huggingface.co/datasets/marwankefah/TriALS/}. To enable independent verification of the registration and label-fusion pipeline, all raw multi-phase CT volumes and per-phase annotations are released alongside the combined labels. The Egyptian data (FWA~000017585) and Chinese data (Sun Yat-Sen IRB) were collected under institutional ethical approval. The external test set labels remain embargoed under the contributing institution's data-sharing agreement; access can be requested through the corresponding authors for verification purposes.

\section*{Code Availability}
All evaluation code, Docker baselines, and scripts are available at \url{https://github.com/xmed-lab/TriALS/tree/main/eval\_scripts}. Pre-trained model weights for the top-ranked methods are publicly available through the challenge repository.

\section*{Acknowledgements}
Supported by the Research Grants Council of Hong Kong (T45-401/22-N), National Natural Science Foundation of China (62306254; 82001768). TriALS is endorsed by SIG-AFRICAI (MICCAI Society).

\section*{Author Contributions}

M.E. and M.G. conceived and designed the challenge. M.E., M.G. and J.M. coordinated data curation and the annotation pipeline at the Egyptian and Chinese training sites. Z.L. led external test cohort data collection at Nanfang Hospital. M.E. performed multi-phase image registration, STAPLE-based label fusion, and constructed the combined-paradigm annotations. M.E. developed the evaluation framework and operated the leaderboard, with leaderboard evaluation support from Y.Q., X.H. and H.W. M.E. designed and analysed the inter-reader study. Z.M. and R.S. served as challenge co-organizers. A.Y. supervised the Egyptian annotation directory.
Mo.G., K.O.N., Ma.E., M.D., A.M.A.S., Nou.E., Noh.E., A.Ab., Am.A., S.A.A., W.D., B.L., H.H. and J.S. performed lesion annotation and provided radiologist verification across multi-phase CT volumes.
Participating challenge teams developed and submitted their respective method entries and reviewed the manuscript: K.E., A.M.M., J.D., M.R., C.U., K.H.M.-H. and F.I. (MIC); S.-W.Y. (WoodyLoveAI); A.E.S. and S.A. (Eagles); X.F., J.Z., X.W., X.M. and G.L. (PCLab-HIT); J.C.D. and J.M.R.S.T. (SliceNDice); A.De., A.Du., Y.X. and I.R. (GenMI); D.K. and M.C. (MedAI); H.Z., Mi.Z. and X.Y. (IMR); A.Q., S.A.N. and Mo.M. (CEMRG); R.E.H., R.M.A., R.M. and X.L.l. (ViCOROB); T.M. and L.E.A. (Caladan); E.A.-M. and A.Ma. (PEDRETS); K.M., E.E.B. and S.K. (BIGR).
X.M.L. supervised the project and acquired funding. M.E. drafted the manuscript and designed the figures. All authors reviewed and approved the final manuscript.

\section*{Competing Interests}
K.H.M.-H.\ and F.I.\ are developers of nnU-Net, which served as the backbone for the majority of participating teams. All other authors declare no competing interests.

\thispagestyle{empty}
\bibliography{mybib}

@misc{rokuss2025voxtell,
      title={VoxTell: Free-Text Promptable Universal 3D Medical Image Segmentation}, 
      author={Maximilian Rokuss and Moritz Langenberg and Yannick Kirchhoff and Fabian Isensee and Benjamin Hamm and Constantin Ulrich and Sebastian Regnery and Lukas Bauer and Efthimios Katsigiannopulos and Tobias Norajitra and Klaus Maier-Hein},
      year={2025},
      eprint={2511.11450},
      archivePrefix={arXiv},
      primaryClass={cs.CV},
      url={https://arxiv.org/abs/2511.11450}, 
}

@misc{lei2026syntheticdatadrivenradiologyfoundation_pastas,
      title={A Synthetic Data-Driven Radiology Foundation Model for Pan-tumor Clinical Diagnosis}, 
      author={Wenhui Lei and Hanyu Chen and Zitian Zhang and Luyang Luo and Qiong Xiao and Yannian Gu and Peng Gao and Yankai Jiang and Ci Wang and Guangtao Wu and Tongjia Xu and Yingjie Zhang and Pranav Rajpurkar and Xiaofan Zhang and Shaoting Zhang and Zhenning Wang},
      year={2026},
      eprint={2502.06171},
      archivePrefix={arXiv},
      primaryClass={eess.IV},
      url={https://arxiv.org/abs/2502.06171}, 
}

@article{liver_cancer_risk_intro,
  title = {Epidemiological and demographic analysis of liver cancer attributable to modifiable risk factors from 1990 to 2021},
  volume = {15},
  ISSN = {2045-2322},
  url = {http://dx.doi.org/10.1038/s41598-025-02031-w},
  DOI = {10.1038/s41598-025-02031-w},
  number = {1},
  journal = {Scientific Reports},
  publisher = {Springer Science and Business Media LLC},
  author = {Zhao,  Yingwu and Peng,  Xue and Zhong,  Zilan and Pan,  Wenhao and Zheng,  Jiyuan and Tian,  Xiaona and Han,  Xinfeng},
  year = {2025},
  month = jun 
}

@article{contrast_supply_issue,
  title = {The environmental sustainability implications of contrast media supply chain disruptions during the COVID-19 pandemic: A document analysis of international practice guidelines},
  volume = {30},
  ISSN = {1078-8174},
  url = {http://dx.doi.org/10.1016/j.radi.2024.05.017},
  DOI = {10.1016/j.radi.2024.05.017},
  journal = {Radiography},
  publisher = {Elsevier BV},
  author = {Natembeya,  M.C. and Anudjo,  M.N.K. and Ackah,  J.A. and Osei,  M.B. and Akudjedu,  T.N.},
  year = {2024},
  month = jun,
  pages = {43–54}
}

@article{liver_stats_in_egypt_intro,
author = {El-Kassas, Mohamed and Khalifa, Rofida and Kohla, Mohamed and Marwan, Eman and Abdelmalek, Mohamed O. and Shousha, Hend},
title = {Hepatocellular Carcinoma in Egypt in the Post HCV Elimination Era: Changing Aetiology, Surveillance, and Management Pathways},
journal = {Liver International},
volume = {46},
number = {4},
pages = {e70595},
doi = {https://doi.org/10.1111/liv.70595},
url = {https://onlinelibrary.wiley.com/doi/abs/10.1111/liv.70595},
eprint = {https://onlinelibrary.wiley.com/doi/pdf/10.1111/liv.70595},
year = {2026}
}

@article{Lits_benchmark,
title = {The Liver Tumor Segmentation Benchmark (LiTS)},
journal = {Medical Image Analysis},
volume = {84},
pages = {102680},
year = {2023},
issn = {1361-8415},
doi = {https://doi.org/10.1016/j.media.2022.102680},
  author = {Bilic,  Patrick and Christ,  Patrick and Li,  Hongwei Bran and Vorontsov,  Eugene and Ben-Cohen,  Avi and Kaissis,  Georgios and Szeskin,  Adi and Jacobs,  Colin and Mamani,  Gabriel Efrain Humpire and Chartrand,  Gabriel and Loh\"{o}fer,  Fabian and Holch,  Julian Walter and Sommer,  Wieland and Hofmann,  Felix and Hostettler,  Alexandre and Lev-Cohain,  Naama and Drozdzal,  Michal and Amitai,  Michal Marianne and Vivanti,  Refael and Sosna,  Jacob and Ezhov,  Ivan and Sekuboyina,  Anjany and Navarro,  Fernando and Kofler,  Florian and Paetzold,  Johannes C. and Shit,  Suprosanna and Hu,  Xiaobin and Lipková,  Jana and Rempfler,  Markus and Piraud,  Marie and Kirschke,  Jan and Wiestler,  Benedikt and Zhang,  Zhiheng and H\"{u}lsemeyer,  Christian and Beetz,  Marcel and Ettlinger,  Florian and Antonelli,  Michela and Bae,  Woong and Bellver,  Míriam and Bi,  Lei and Chen,  Hao and Chlebus,  Grzegorz and Dam,  Erik B. and Dou,  Qi and Fu,  Chi-Wing and Georgescu,  Bogdan and Giró-i-Nieto,  Xavier and Gruen,  Felix and Han,  Xu and Heng,  Pheng-Ann and Hesser,  J\"{u}rgen and Moltz,  Jan Hendrik and Igel,  Christian and Isensee,  Fabian and J\"{a}ger,  Paul and Jia,  Fucang and Kaluva,  Krishna Chaitanya and Khened,  Mahendra and Kim,  Ildoo and Kim,  Jae-Hun and Kim,  Sungwoong and Kohl,  Simon and Konopczynski,  Tomasz and Kori,  Avinash and Krishnamurthi,  Ganapathy and Li,  Fan and Li,  Hongchao and Li,  Junbo and Li,  Xiaomeng and Lowengrub,  John and Ma,  Jun and Maier-Hein,  Klaus and Maninis,  Kevis-Kokitsi and Meine,  Hans and Merhof,  Dorit and Pai,  Akshay and Perslev,  Mathias and Petersen,  Jens and Pont-Tuset,  Jordi and Qi,  Jin and Qi,  Xiaojuan and Rippel,  Oliver and Roth,  Karsten and Sarasua,  Ignacio and Schenk,  Andrea and Shen,  Zengming and Torres,  Jordi and Wachinger,  Christian and Wang,  Chunliang and Weninger,  Leon and Wu,  Jianrong and Xu,  Daguang and Yang,  Xiaoping and Yu,  Simon Chun-Ho and Yuan,  Yading and Yue,  Miao and Zhang,  Liping and Cardoso,  Jorge and Bakas,  Spyridon and Braren,  Rickmer and Heinemann,  Volker and Pal,  Christopher and Tang,  An and Kadoury,  Samuel and Soler,  Luc and van Ginneken,  Bram and Greenspan,  Hayit and Joskowicz,  Leo and Menze,  Bjoern},
url = {https://www.sciencedirect.com/science/article/pii/S1361841522003085},
}

@article{huang2023stunet,
  title={STU-Net: Scalable and Transferable Medical Image Segmentation Models Empowered by Large-Scale Supervised Pre-training},
  author={Huang, Ziyan and Wang, Haoyu and Deng, Zhongying and Ye, Jin and Su, Yanzhou and Sun, Hui and He, Junjun and Gu, Yun and Gu, Lixu and Zhang, Shaoting and Qiao, Yu},
  journal={arXiv preprint arXiv:2304.06716},
  year={2023}
}

@misc{EgyptIndependent2025_CTContrastShortage,
  author       = {{Egypt Independent}},
  title        = {Dye shortage in {Egypt} for {CT} scans threatens patient lives},
  year         = {2025},
  url          = {https://www.egyptindependent.com/dye-shortage-in-egypt-for-ct-scans-threatens-patient-lives/},
  note         = {Accessed: 2025-06-01}
}

@misc{heller2023kits21challengeautomaticsegmentation,
      title={The KiTS21 Challenge: Automatic segmentation of kidneys, renal tumors, and renal cysts in corticomedullary-phase CT}, 
      author={Nicholas Heller and Fabian Isensee and Dasha Trofimova and Resha Tejpaul and Zhongchen Zhao and Huai Chen and Lisheng Wang and Alex Golts and Daniel Khapun and Daniel Shats and Yoel Shoshan and Flora Gilboa-Solomon and Yasmeen George and Xi Yang and Jianpeng Zhang and Jing Zhang and Yong Xia and Mengran Wu and Zhiyang Liu and Ed Walczak and Sean McSweeney and Ranveer Vasdev and Chris Hornung and Rafat Solaiman and Jamee Schoephoerster and Bailey Abernathy and David Wu and Safa Abdulkadir and Ben Byun and Justice Spriggs and Griffin Struyk and Alexandra Austin and Ben Simpson and Michael Hagstrom and Sierra Virnig and John French and Nitin Venkatesh and Sarah Chan and Keenan Moore and Anna Jacobsen and Susan Austin and Mark Austin and Subodh Regmi and Nikolaos Papanikolopoulos and Christopher Weight},
      year={2023},
      eprint={2307.01984},
      archivePrefix={arXiv},
      primaryClass={cs.CV},
      url={https://arxiv.org/abs/2307.01984}, 
}

@article{Wu2025,
  title = {A Multi-phase CT Dataset for Automated Differential Diagnosis of Liver Tumors},
  volume = {13},
  ISSN = {2052-4463},
  url = {http://dx.doi.org/10.1038/s41597-025-06343-4},
  DOI = {10.1038/s41597-025-06343-4},
  number = {1},
  journal = {Scientific Data},
  publisher = {Springer Science and Business Media LLC},
  author = {Wu,  Xiang´an and Su,  Haoyang and Hua,  Yiwei and Xu,  Yali and Wang,  Lilong and Wang,  Xiaosong and Wang,  Shitian and Jin,  Bao and Liu,  Xiao and Wan,  Xueshuai and Sun,  Qiang and Wang,  Xuan and Du,  Shunda},
  year = {2025},
  month = dec 
}

@misc{ji2022amoslargescaleabdominalmultiorgan,
      title={AMOS: A Large-Scale Abdominal Multi-Organ Benchmark for Versatile Medical Image Segmentation}, 
      author={Yuanfeng Ji and Haotian Bai and Jie Yang and Chongjian Ge and Ye Zhu and Ruimao Zhang and Zhen Li and Lingyan Zhang and Wanling Ma and Xiang Wan and Ping Luo},
      year={2022},
      eprint={2206.08023},
      archivePrefix={arXiv},
      primaryClass={eess.IV},
      url={https://arxiv.org/abs/2206.08023}, 
}

@inbook{ulrich2023multitalent,
  title = {MultiTalent: A Multi-dataset Approach to Medical Image Segmentation},
  ISBN = {9783031438981},
  ISSN = {1611-3349},
  url = {http://dx.doi.org/10.1007/978-3-031-43898-1_62},
  DOI = {10.1007/978-3-031-43898-1_62},
  booktitle = {Medical Image Computing and Computer Assisted Intervention – MICCAI 2023},
  publisher = {Springer Nature Switzerland},
  author = {Ulrich,  Constantin and Isensee,  Fabian and Wald,  Tassilo and Zenk,  Maximilian and Baumgartner,  Michael and Maier-Hein,  Klaus H.},
  year = {2023},
  pages = {648–658}
}

@article{isensee2021nnunet,
  title = {nnU-Net: a self-configuring method for deep learning-based biomedical image segmentation},
  volume = {18},
  ISSN = {1548-7105},
  url = {http://dx.doi.org/10.1038/s41592-020-01008-z},
  DOI = {10.1038/s41592-020-01008-z},
  number = {2},
  journal = {Nature Methods},
  publisher = {Springer Science and Business Media LLC},
  author = {Isensee,  Fabian and Jaeger,  Paul F. and Kohl,  Simon A. A. and Petersen,  Jens and Maier-Hein,  Klaus H.},
  year = {2020},
  month = dec,
  pages = {203–211}
}

@inbook{roy2023mednext,
  title = {MedNeXt: Transformer-Driven Scaling of ConvNets for Medical Image Segmentation},
  ISBN = {9783031439018},
  ISSN = {1611-3349},
  url = {http://dx.doi.org/10.1007/978-3-031-43901-8_39},
  DOI = {10.1007/978-3-031-43901-8_39},
  booktitle = {Medical Image Computing and Computer Assisted Intervention – MICCAI 2023},
  publisher = {Springer Nature Switzerland},
  author = {Roy,  Saikat and Koehler,  Gregor and Ulrich,  Constantin and Baumgartner,  Michael and Petersen,  Jens and Isensee,  Fabian and J\"{a}ger,  Paul F. and Maier-Hein,  Klaus H.},
  year = {2023},
  pages = {405–415}
}

@article{maier-hein_2018_bias,
  title = {Why rankings of biomedical image analysis competitions should be interpreted with care},
  volume = {9},
  ISSN = {2041-1723},
  url = {http://dx.doi.org/10.1038/s41467-018-07619-7},
  DOI = {10.1038/s41467-018-07619-7},
  number = {1},
  journal = {Nature Communications},
  publisher = {Springer Science and Business Media LLC},
  author = {Maier-Hein,  Lena and Eisenmann,  Matthias and Reinke,  Annika and Onogur,  Sinan and Stankovic,  Marko and Scholz,  Patrick and Arbel,  Tal and Bogunovic,  Hrvoje and Bradley,  Andrew P. and Carass,  Aaron and Feldmann,  Carolin and Frangi,  Alejandro F. and Full,  Peter M. and van Ginneken,  Bram and Hanbury,  Allan and Honauer,  Katrin and Kozubek,  Michal and Landman,  Bennett A. and M\"{a}rz,  Keno and Maier,  Oskar and Maier-Hein,  Klaus and Menze,  Bjoern H. and M\"{u}ller,  Henning and Neher,  Peter F. and Niessen,  Wiro and Rajpoot,  Nasir and Sharp,  Gregory C. and Sirinukunwattana,  Korsuk and Speidel,  Stefanie and Stock,  Christian and Stoyanov,  Danail and Taha,  Abdel Aziz and van der Sommen,  Fons and Wang,  Ching-Wei and Weber,  Marc-André and Zheng,  Guoyan and Jannin,  Pierre and Kopp-Schneider,  Annette},
  year = {2018},
  month = dec 
}

@ARTICLE{elastix,
  author={Klein, Stefan and Staring, Marius and Murphy, Keelin and Viergever, Max A. and Pluim, Josien P. W.},
  journal={IEEE Transactions on Medical Imaging}, 
  title={elastix: A Toolbox for Intensity-Based Medical Image Registration}, 
  year={2010},
  volume={29},
  number={1},
  pages={196-205},
  doi={10.1109/TMI.2009.2035616}}

@article{Reinke2024,
  title = {Understanding metric-related pitfalls in image analysis validation},
  volume = {21},
  ISSN = {1548-7105},
  url = {http://dx.doi.org/10.1038/s41592-023-02150-0},
  DOI = {10.1038/s41592-023-02150-0},
  number = {2},
  journal = {Nature Methods},
  publisher = {Springer Science and Business Media LLC},
  author = {Reinke,  Annika and Tizabi,  Minu D. and Baumgartner,  Michael and Eisenmann,  Matthias and Heckmann-N\"{o}tzel,  Doreen and Kavur,  A. Emre and R\"{a}dsch,  Tim and Sudre,  Carole H. and Acion,  Laura and Antonelli,  Michela and Arbel,  Tal and Bakas,  Spyridon and Benis,  Arriel and Buettner,  Florian and Cardoso,  M. Jorge and Cheplygina,  Veronika and Chen,  Jianxu and Christodoulou,  Evangelia and Cimini,  Beth A. and Farahani,  Keyvan and Ferrer,  Luciana and Galdran,  Adrian and van Ginneken,  Bram and Glocker,  Ben and Godau,  Patrick and Hashimoto,  Daniel A. and Hoffman,  Michael M. and Huisman,  Merel and Isensee,  Fabian and Jannin,  Pierre and Kahn,  Charles E. and Kainmueller,  Dagmar and Kainz,  Bernhard and Karargyris,  Alexandros and Kleesiek,  Jens and Kofler,  Florian and Kooi,  Thijs and Kopp-Schneider,  Annette and Kozubek,  Michal and Kreshuk,  Anna and Kurc,  Tahsin and Landman,  Bennett A. and Litjens,  Geert and Madani,  Amin and Maier-Hein,  Klaus and Martel,  Anne L. and Meijering,  Erik and Menze,  Bjoern and Moons,  Karel G. M. and M\"{u}ller,  Henning and Nichyporuk,  Brennan and Nickel,  Felix and Petersen,  Jens and Rafelski,  Susanne M. and Rajpoot,  Nasir and Reyes,  Mauricio and Riegler,  Michael A. and Rieke,  Nicola and Saez-Rodriguez,  Julio and Sánchez,  Clara I. and Shetty,  Shravya and Summers,  Ronald M. and Taha,  Abdel A. and Tiulpin,  Aleksei and Tsaftaris,  Sotirios A. and Van Calster,  Ben and Varoquaux,  Gaël and Yaniv,  Ziv R. and J\"{a}ger,  Paul F. and Maier-Hein,  Lena},
  year = {2024},
  month = feb,
  pages = {182–194}
}

@article{Warfield2004,
  title = {Simultaneous Truth and Performance Level Estimation (STAPLE): An Algorithm for the Validation of Image Segmentation},
  volume = {23},
  ISSN = {0278-0062},
  url = {http://dx.doi.org/10.1109/TMI.2004.828354},
  DOI = {10.1109/tmi.2004.828354},
  number = {7},
  journal = {IEEE Transactions on Medical Imaging},
  publisher = {Institute of Electrical and Electronics Engineers (IEEE)},
  author = {Warfield,  S.K. and Zou,  K.H. and Wells,  W.M.},
  year = {2004},
  month = july,
  pages = {903–921}
}

@article{MAIERHEIN2020101796,
title = {BIAS: Transparent reporting of biomedical image analysis challenges},
journal = {Medical Image Analysis},
volume = {66},
pages = {101796},
year = {2020},
issn = {1361-8415},
doi = {https://doi.org/10.1016/j.media.2020.101796},
url = {https://www.sciencedirect.com/science/article/pii/S1361841520301602},
author = {Lena Maier-Hein and Annika Reinke and Michal Kozubek and Anne L. Martel and Tal Arbel and Matthias Eisenmann and Allan Hanbury and Pierre Jannin and Henning Müller and Sinan Onogur and Julio Saez-Rodriguez and Bram {van Ginneken} and Annette Kopp-Schneider and Bennett A. Landman}
}

@article{fdr,
 ISSN = {00359246},
 URL = {http://www.jstor.org/stable/2346101},
 author = {Yoav Benjamini and Yosef Hochberg},
 journal = {Journal of the Royal Statistical Society. Series B (Methodological)},
 number = {1},
 pages = {289--300},
 publisher = {[Royal Statistical Society, Oxford University Press]},
 title = {Controlling the False Discovery Rate: A Practical and Powerful Approach to Multiple Testing},
 urldate = {2026-04-20},
 volume = {57},
 year = {1995}
}

@InProceedings{SAM,
    author    = {Kirillov, Alexander and Mintun, Eric and Ravi, Nikhila and Mao, Hanzi and Rolland, Chloe and Gustafson, Laura and Xiao, Tete and Whitehead, Spencer and Berg, Alexander C. and Lo, Wan-Yen and Dollar, Piotr and Girshick, Ross},
    title     = {Segment Anything},
    booktitle = {Proceedings of the IEEE/CVF International Conference on Computer Vision (ICCV)},
    month     = {October},
    year      = {2023},
    pages     = {4015-4026}
}

@article{interactive,
  author       = {Tomas Sakinis and
                  Fausto Milletari and
                  Holger Roth and
                  Panagiotis Korfiatis and
                  Petro M. Kostandy and
                  Kenneth Philbrick and
                  Zeynettin Akkus and
                  Ziyue Xu and
                  Daguang Xu and
                  Bradley J. Erickson},
  title        = {Interactive segmentation of medical images through fully convolutional
                  neural networks},
  journal      = {CoRR},
  volume       = {abs/1903.08205},
  year         = {2019},
  url          = {http://arxiv.org/abs/1903.08205},
  eprinttype   = {arXiv},
  eprint       = {1903.08205},
  bibsource    = {dblp computer science bibliography, https://dblp.org}
}

@article{Ma2024,
  title = {Segment anything in medical images},
  volume = {15},
  ISSN = {2041-1723},
  url = {http://dx.doi.org/10.1038/s41467-024-44824-z},
  DOI = {10.1038/s41467-024-44824-z},
  number = {1},
  journal = {Nature Communications},
  publisher = {Springer Science and Business Media LLC},
  author = {Ma,  Jun and He,  Yuting and Li,  Feifei and Han,  Lin and You,  Chenyu and Wang,  Bo},
  year = {2024},
  month = jan 
}

\clearpage

\clearpage


\begin{table}[htbp]
\centering
\caption{\textbf{TriALS dataset composition, lesion statistics, and CT acquisition parameters.}
Lesion counts and volumes are from the venous phase.
Acquisition parameters are from the non-contrast phase (mean$\pm$SD).}
\label{tab:cohort}
\small
\setlength{\tabcolsep}{3pt}
\begin{tabular}{llccccccccc}
\toprule
& & & & \multicolumn{4}{c}{\textbf{Lesion statistics (venous)}} & \multicolumn{3}{c}{\textbf{CT acquisition (NC)}}\\
\cmidrule(lr){5-8}\cmidrule(lr){9-11}
\textbf{Institution} & \textbf{Country} & \textbf{Split} & \textbf{Cases} & \textbf{Vols} & \textbf{Lesions} & \textbf{Med.\ vol.} & \textbf{IQR (mm$^3$)} & \textbf{In-plane} & \textbf{Slice} & \textbf{Depth}\\
\midrule
Ain Shams Univ.   & Egypt  & Train & 60 & 240 & \multirow{2}{*}{175} & \multirow{2}{*}{1{,}573} & \multirow{2}{*}{308--7{,}908} & $0.854\pm0.099$ & $0.97\pm0.19$ & $464\pm175$ \\
Sun Yat-Sen Mem.  & China  & Train & 20 & 80  &     &        &                & $0.766\pm0.063$ & $0.98\pm0.12$ & $391\pm186$ \\
\midrule
Ain Shams Univ.   & Egypt  & Test  & 28 & 112 & 222 & 556    & 151--1{,}813   & $0.862\pm0.095$ & $0.86\pm0.15$ & $578\pm137$ \\
Sun Yat-Sen Mem.  & China  & Test  & 17 & 68  & 31  & 10{,}658 & 554--117{,}859 & $0.766\pm0.063$ & $0.98\pm0.12$ & $391\pm186$ \\
Nanfang Hospital  & China  & Test  & 25 & 100 & 82  & 604    & 90--42{,}752   & $0.686\pm0.072$ & $1.02\pm0.10$ & $240\pm69$  \\
\midrule
\textbf{Total}    &        &       & \textbf{150} & \textbf{600} & \textbf{510} & & & & & \\
\bottomrule
\end{tabular}
\end{table}

\begin{table}[htbp]
\centering
\caption{\textbf{Pairwise centre comparisons: acquisition parameters and segmentation metrics.}
Median values with interquartile range [IQR] are reported per test centre.
$p$-values were obtained by Mann–Whitney $U$ test (two-sided); an alpha level of 0.05
was used to determine significance after Benjamini--Hochberg FDR correction within each parameter family. Bold values indicate $p < 0.05$ (FDR-corrected).
Acquisition statistics are from the non-contrast phase.
Segmentation metrics are from TriALS~2025 across all participating teams.}
\label{tab:centre_pvalues}
\small
\setlength{\tabcolsep}{4pt}
\begin{tabular}{lcccccc}
\toprule
& \multicolumn{3}{c}{\textbf{Median [IQR]}} & \multicolumn{3}{c}{\textbf{$p$-value}} \\
\cmidrule(lr){2-4}\cmidrule(lr){5-7}
\textbf{Parameter} & \textbf{China} & \textbf{Egypt} & \textbf{External}
                   & \textbf{C vs E} & \textbf{C vs X} & \textbf{E vs X} \\
\midrule
\multicolumn{7}{l}{\textit{CT acquisition (non-contrast phase)}} \\
In-plane (mm)       & 0.768 [0.719--0.824] & 0.839 [0.784--0.902] & 0.691 [0.644--0.754] & \textbf{0.004} & \textbf{0.003} & \textbf{$<$0.001} \\
Slice thickness (mm) & 1.00 [1.00--1.00]   & 0.80 [0.80--0.80]   & 1.00 [1.00--1.00]   & \textbf{0.006} & 0.096          & \textbf{$<$0.001} \\
Depth (slices)       & 426 [361--525]       & 610 [526--643]       & 226 [214--236]       & \textbf{0.002} & \textbf{$<$0.001} & \textbf{$<$0.001} \\
\midrule
\multicolumn{7}{l}{\textit{Task 1 (venous-phase CT)}} \\
Dice/case           & 0.542 [0.436--0.557] & 0.620 [0.576--0.640] & 0.591 [0.535--0.630] & 0.209 & 0.259 & 0.535 \\
ASSD ($\times10^3$ mm) & 8.47 [8.33--8.82]  & 6.69 [6.54--7.22]   & 7.83 [7.74--8.34]   & \textbf{0.007} & 0.038$^\dagger$ & 0.017$^\dagger$ \\
\midrule
\multicolumn{7}{l}{\textit{Task 2 (non-contrast CT)}} \\
Dice/case (visible)  & 0.494 [0.425--0.510] & 0.362 [0.260--0.416] & 0.527 [0.506--0.556] & 0.180 & 0.240 & 0.065 \\
Dice/case (combined) & 0.460 [0.390--0.493] & 0.294 [0.238--0.350] & 0.511 [0.479--0.540] & 0.065 & 0.240 & \textbf{0.004} \\
\bottomrule
\end{tabular}
\begin{tablenotes}
\small
\item C\,=\,China; E\,=\,Egypt; X\,=\,External. Acquisition $p$-values are per-case (NC phase volumes).
Segmentation $p$-values compare the distribution of team-level scores across centres
($n\!=\!7$ teams Task~1; $n\!=\!6$ teams Task~2). Bold indicates $p < 0.05$ after Benjamini--Hochberg FDR correction within each parameter family. $^\dagger$Nominally significant ($p < 0.05$) but not significant after FDR correction.
\end{tablenotes}
\end{table}

\begin{table}[htbp]
\centering
\caption{\textbf{TriALS 2025 Task~1 results across three centres (7 teams).} Per-case Dice and ASSD ($\times10^3$\,mm). ASSD values include a fixed penalty of $9{,}001$\,mm for each undetected (false-negative) and spurious (false-positive) lesion, aggregated across all lesions per case; values therefore reflect both segmentation accuracy and detection completeness (see Methods for penalty convention).}
\label{tab:trials2025_t1}
\small
\begin{tabular}{lcccccc}
\toprule
& \multicolumn{2}{c}{\textbf{China}} & \multicolumn{2}{c}{\textbf{Egypt}} & \multicolumn{2}{c}{\textbf{External}}\\
\cmidrule(lr){2-3}\cmidrule(lr){4-5}\cmidrule(lr){6-7}
\textbf{Team} & Dice/case & ASSD & Dice/case & ASSD & Dice/case & ASSD\\
\midrule
mic        & \textbf{0.721} & 8.05 & \textbf{0.791} & 6.33 & \textbf{0.750} & 7.62\\
hit        & 0.556 & 8.43 & 0.608 & 6.69 & 0.665 & 7.83\\
wlai       & 0.557 & 8.24 & 0.620 & 6.61 & 0.595 & 7.75\\
eagles     & 0.542 & 8.47 & 0.637 & 7.00 & 0.591 & 7.73\\
gen\_bold  & 0.376 & 8.80 & 0.644 & 6.47 & 0.535 & 8.25\\
slicendice & 0.447 & 8.83 & 0.543 & 7.44 & 0.535 & 8.43\\
caladan    & 0.425 & 8.84 & 0.280 & 8.53 & 0.464 & 8.69\\
\midrule
\textit{Mean} & \textit{0.518} & — & \textit{0.589} & — & \textit{0.591} & —\\
\bottomrule
\end{tabular}
\end{table}

\begin{table}[htbp]
\centering
\caption{\textbf{TriALS 2025 Task~2 (non-contrast CT) results across three centres (6 teams).}}
\label{tab:trials2025_t2}
\small
\begin{tabular}{lcccccc}
\toprule
& \multicolumn{2}{c}{\textbf{China}} & \multicolumn{2}{c}{\textbf{Egypt}} & \multicolumn{2}{c}{\textbf{External}}\\
\cmidrule(lr){2-3}\cmidrule(lr){4-5}\cmidrule(lr){6-7}
\textbf{Team} & Combined & Visible & Combined & Visible & Combined & Visible\\
\midrule
mic       & \textbf{0.586} & \textbf{0.621} & \textbf{0.471} & \textbf{0.552} & \textbf{0.651} & \textbf{0.674}\\
eagles    & 0.502 & 0.513 & 0.323 & 0.350 & 0.539 & 0.537\\
wlai      & 0.466 & 0.488 & 0.204 & 0.230 & 0.483 & 0.516\\
medai     & 0.454 & 0.501 & 0.359 & 0.430 & 0.477 & 0.502\\
hit       & 0.368 & 0.404 & 0.264 & 0.374 & 0.541 & 0.563\\
caladan   & 0.321 & 0.286 & 0.147 & 0.101 & 0.373 & 0.352\\
\midrule
\textit{Mean} & \textit{0.449} & \textit{0.469} & \textit{0.295} & \textit{0.340} & \textit{0.511} & \textit{0.524}\\
\bottomrule
\end{tabular}
\end{table}

\begin{table}[htbp]
\centering
\caption{\textbf{TriALS 2024 results (Egyptian test set; 6 teams).} Sorted by Task~1 per-case Dice.}
\label{tab:trials2024}
\small
\begin{tabular}{lcccccc}
\toprule
& \multicolumn{2}{c}{\textbf{Task~1 (Venous)}} & \multicolumn{2}{c}{\textbf{Task~2 Combined}} & \multicolumn{2}{c}{\textbf{Task~2 Visible}}\\
\cmidrule(lr){2-3}\cmidrule(lr){4-5}\cmidrule(lr){6-7}
\textbf{Team} & Dice/case & Dice/glob & Dice/case & Dice/glob & Dice/case & Dice/glob\\
\midrule
mic        & \textbf{0.742} & \textbf{0.824} & \textbf{0.429} & \textbf{0.410} & \textbf{0.553} & \textbf{0.400}\\
vicorob    & 0.698 & 0.667 & 0.427 & 0.403 & 0.433 & 0.345\\
imr        & 0.666 & 0.771 & 0.385 & 0.351 & 0.459 & 0.330\\
cemrg      & 0.576 & 0.524 & 0.270 & 0.303 & 0.338 & 0.249\\
bigr       & 0.547 & 0.652 & 0.303 & 0.401 & 0.315 & 0.354\\
pedrets    & 0.496 & 0.659 & 0.372 & 0.391 & 0.414 & 0.349\\
\midrule
\textit{Mean} & \textit{0.622} & — & \textit{0.364} & — & \textit{0.419} & —\\
\bottomrule
\end{tabular}
\end{table}

\clearpage
\setcounter{figure}{0}
\setcounter{table}{0}
\renewcommand{\thefigure}{S\arabic{figure}}
\renewcommand{\thetable}{S\arabic{table}}

\section*{Supplementary Figures}
\begin{figure}[htbp]
  \centering
  \includegraphics[width=\linewidth]{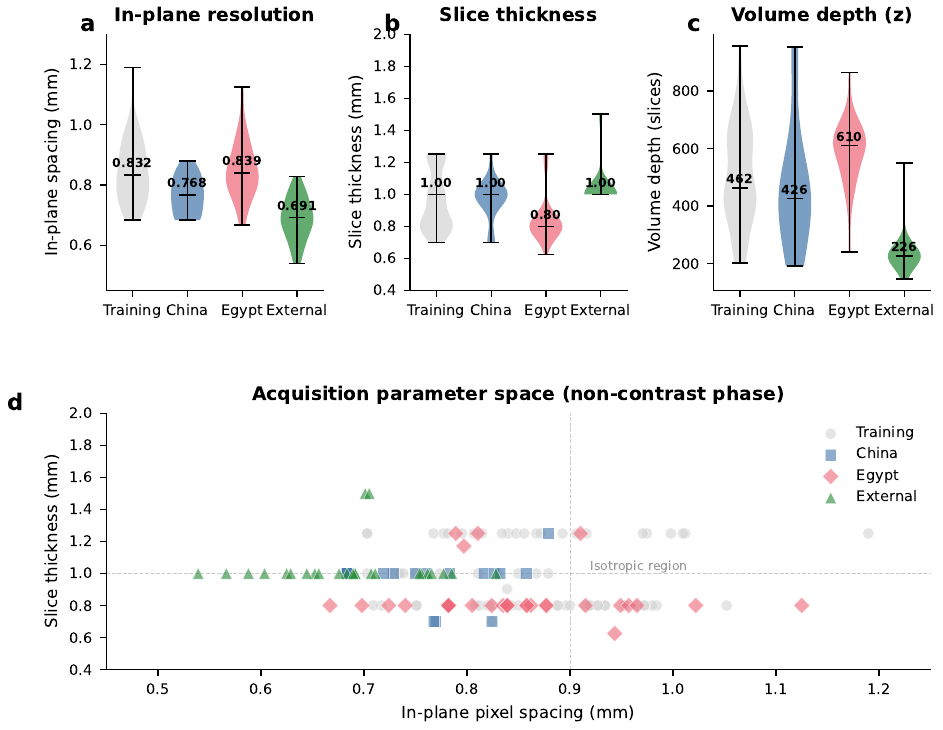}
  \caption{\textbf{CT acquisition parameter heterogeneity across cohorts.}
    \textbf{a,}~In-plane pixel spacing distributions (non-contrast phase). \textbf{b,}~Slice thickness distributions. \textbf{c,}~Volume depth (number of axial slices).
    \textbf{d,}~Joint distribution of in-plane spacing and slice thickness; each point represents one non-contrast volume.
    Dashed lines mark isotropic resolution boundaries. Significant differences in all three parameters were confirmed across centre pairs (Mann--Whitney $U$, $p < 0.05$ for 8 of 9 comparisons; Table~\ref{tab:centre_pvalues}).}
  \label{fig:spacing}
\end{figure}

\begin{figure}[htbp]
  \centering
  \includegraphics[width=\linewidth]{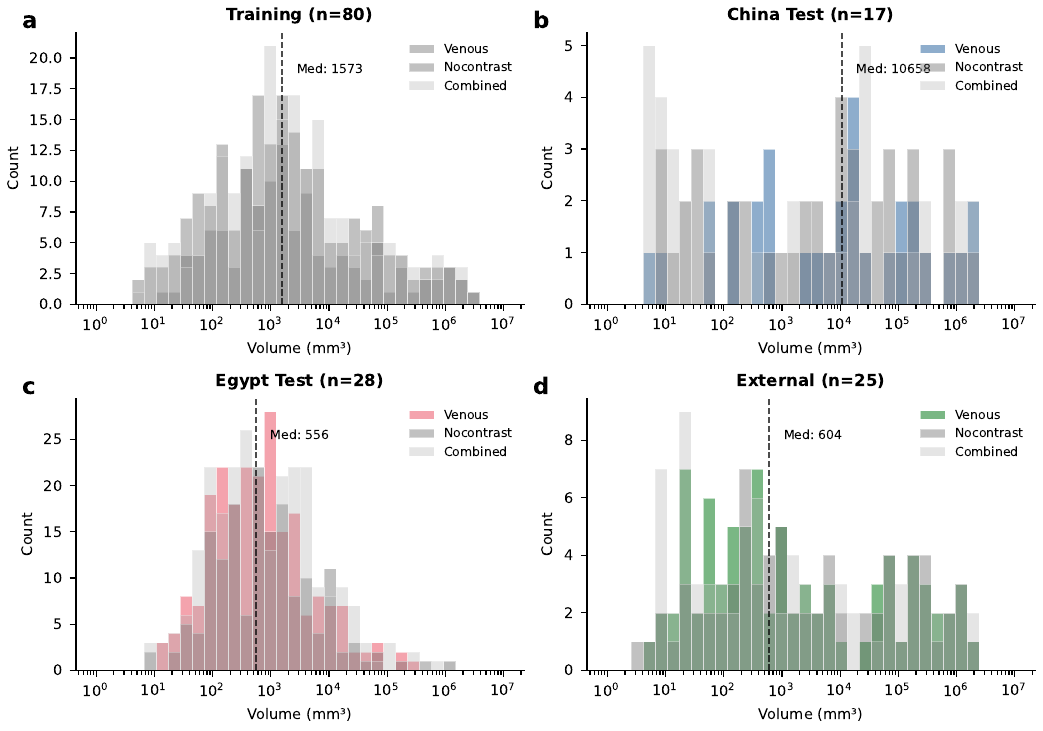}
  \caption{\textbf{Lesion volume distributions by cohort and phase.}
    Histograms (log$_{10}$ scale) for \textbf{a,}~training, \textbf{b,}~China test, \textbf{c,}~Egypt test, \textbf{d,}~external. Dashed lines mark venous-phase median volumes.}
  \label{edfig:lesion_size}
\end{figure}

\begin{figure}[htbp]
  \centering
  \includegraphics[width=\linewidth]{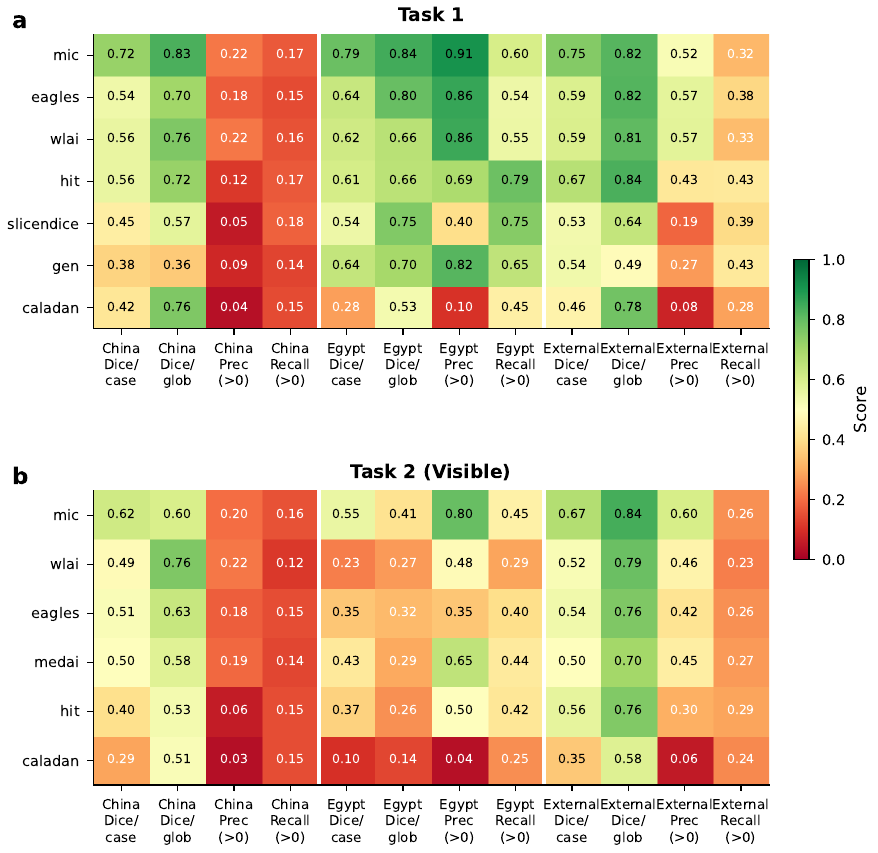}
  \caption{\textbf{Multi-metric performance heatmap.}
    Green = high; red = low. \textbf{a,}~Task~1; \textbf{b,}~Task~2 visible. Columns = metric $\times$ centre.}
  \label{edfig:heatmap}
\end{figure}
\clearpage
\section*{Supplementary Tables}

\begin{table}[!htbp]
\centering
\caption{\textbf{Lesion size distribution (venous phase).}
Size categories in mm$^3$.
Percentages in parentheses.}
\label{supp:tab:lesion_size}
\small
\begin{tabular}{@{}lc C{1.1cm}C{1.1cm}C{1.1cm}C{1.2cm}C{1.1cm} r@{}}
\toprule
\textbf{Cohort} & $N$ & $<$100 & 100--1k & 1k--10k & 10k--100k & $>$100k & \textbf{Median} \\
\midrule
Training   & 175 & 21 (12\%) & 57 (33\%) & 56 (32\%) & 23 (13\%) & 18 (10\%) & 1{,}573 \\
China test & 31  & 4 (13\%)  & 7 (23\%)  & 4 (13\%)  & 7 (23\%)  & 9 (29\%)  & 10{,}658 \\
Egypt test & 222 & 37 (17\%) & 105 (47\%)& 59 (27\%) & 18 (8\%)  & 3 (1\%)   & 556 \\
External   & 82  & 22 (27\%) & 25 (30\%) & 11 (13\%) & 8 (10\%)  & 16 (20\%) & 604 \\
\bottomrule
\end{tabular}
\end{table}

\begin{table}[!htbp]
\centering
\caption{\textbf{Inter-reader agreement between two radiologists across CT phases.}
Per-case Dice (mean $\pm$ SD), global Dice, lesion-level precision and recall (detection threshold: $\geq$10\% IoU).
DSC\,=\,0 is assigned for lesions annotated by one reader but missed by the other, matching the penalty convention used for automated methods.}
\label{tab:reader}
\small
\begin{tabular}{@{}lccccc@{}}
\toprule
\textbf{Phase} & \textbf{Per-case Dice} & \textbf{Global Dice} & \textbf{Precision} & \textbf{Recall} \\
\midrule
Venous        & $0.742\pm0.443$ & 0.914 & 0.833 & 0.946 \\
Non-contrast  & $0.574\pm0.490$ & 0.692 & 0.333 & 0.769 \\
\bottomrule
\end{tabular}
\end{table}

\begin{table}[!htbp]
\centering
\caption{\textbf{Performance comparison: TriALS~2025 top methods vs.\ off-the-shelf state-of-the-art.}
Per-case Dice on Task~1 (venous CT) and Task~2 combined (non-contrast CT) across three test centres.
Inter-reader (IR) ceilings: venous 0.742, non-contrast 0.574.
Bold = best per column.}
\label{tab:sota}
\small
\begin{tabular}{@{}llcccccc@{}}
\toprule
& & \multicolumn{3}{c}{\textbf{Task 1 --- Venous (D/case)}} & \multicolumn{3}{c}{\textbf{Task 2 Combined --- NCCT (D/case)}} \\
\cmidrule(lr){3-5} \cmidrule(lr){6-8}
\textbf{Method} & \textbf{Type} & China & Egypt & Ext. & China & Egypt & Ext. \\
\midrule
\textbf{mic\_v2} (rank~1)  & Challenge  & \textbf{.721} & \textbf{.791} & \textbf{.750} & \textbf{.586} & \textbf{.471} & \textbf{.651} \\
VoxTell~1.1        & Off-the-shelf & .656 & .650 & .686 & .382 & .330 & .553 \\
PASTA              & Off-the-shelf & .499 & .700 & .513 & .289 & .290 & .369 \\
\addlinespace
IR ceiling         & Human         & \multicolumn{3}{c}{0.742 (venous)} & \multicolumn{3}{c}{0.574 (non-contrast)} \\
\bottomrule
\end{tabular}
\end{table}

\begin{table}[!htbp]
\centering
\caption{\textbf{Summary of participating methods.}
Architecture, training strategy, and key design choices for all teams across both challenge editions.
T1 = Task~1 (venous-phase); NC = non-contrast CT task.
Ext.\ data = external training data beyond TriALS.
Ens.\ = ensemble at inference.
}
\label{tab:supp_methods}
\scriptsize
\renewcommand{\arraystretch}{1.25}
\begin{tabular}{@{} L{1.1cm} L{1.4cm} L{1.7cm} L{1.9cm} L{1.4cm} L{0.9cm} L{1.0cm} C{0.4cm} C{0.4cm} @{}}
\toprule
\textbf{Year} & \textbf{Team} & \textbf{Architecture} & \textbf{Pre-training} & \textbf{Ext.\ data} & \textbf{Epochs} & \textbf{Loss} & \textbf{T1} & \textbf{NC} \\
\midrule
\multirow{6}{*}{2024}
 & MIC         & ResEnc-L U-Net    & MultiTalent (30 datasets) & LiTS + multi-phase  & 200    & Dice+CE  & \checkmark & \checkmark \\
 & ViCOROB     & PlainConv/ResEnc  & LiTS pre-train            & LiTS                & 1000   & Dice+CE  & \checkmark & \checkmark \\
 & IMR         & PlainConv U-Net   & Liver seg.\ (4 datasets)  & SLIVER07, LiTS, CHAOS, AMOS & 1000 & Dice+CE  & \checkmark & \checkmark \\
 & CEMRG       & xLSTM-UNet        & SSL student--teacher       & None                & N/A    & Contrastive+Dice+CE & \checkmark & \checkmark \\
 & PEDRETS     & MedNeXt-L (5$\times$5) & LiTS pre-train        & LiTS                & 100+1000 & Dice+CE & \checkmark & \checkmark \\
 & BIGR        & MedNeXt-L (5$\times$5) & None                  & None                & 100/300  & Dice+CE & ---        & \checkmark \\
\addlinespace
\midrule
\multirow{8}{*}{2025}
 & MIC         & ResEnc-L U-Net    & MultiTalent (86 datasets)  & LiTS, HCC, LiverMets, HCC-TACE, WAW-TACE & 1050 & Dice+CE & \checkmark & \checkmark \\
 & Eagles      & 3D nnU-Net ResNet-M & None                     & None                & Default  & Dice+CE & \checkmark & \checkmark \\
 & WoodyLoveAI & Custom ResEnc (multi-scale) & MSD Liver $\rightarrow$ CECT & MSD Liver, CECT & 1000 & Dice+CE & \checkmark & \checkmark \\
 & PCLab-HIT   & nnU-Net + contrastive & TriALS baseline        & None                & $\sim$1000 & Dice+CE+InfoNCE & \checkmark & \checkmark \\
 & SlicenDice  & RFA Res-UNet      & None                       & LiTS                & 200    & Dice+CE  & \checkmark & --- \\
 & GenMI       & nnU-Net (plain)   & None                       & None                & 500    & Dice+Focal & \checkmark & --- \\
 & CALADAN     & MedNeXt-M (3$\times$3) & Contrast phases        & None                & 50$\times$3 & Dice+CE & \checkmark & \checkmark \\
 & MedAI       & STU-Net (region)  & None                       & None                & 1000   & Dice+CE (region) & --- & \checkmark \\
\bottomrule
\end{tabular}
\end{table}

\begin{table}[!htbp]
\centering
\caption{\textbf{Technical implementation details.}
Patch size, batch size, optimizer, GPU, and inference strategy for each team.}
\label{tab:supp_technical}
\scriptsize
\renewcommand{\arraystretch}{1.2}
\begin{tabular}{@{} L{1.0cm} L{1.2cm} C{1.8cm} C{0.6cm} L{1.4cm} L{1.6cm} L{1.6cm} @{}}
\toprule
\textbf{Year} & \textbf{Team} & \textbf{Patch size} & \textbf{BS} & \textbf{Optimizer} & \textbf{GPU} & \textbf{Inference} \\
\midrule
\multirow{6}{*}{2024}
 & MIC      & 192$^3$          & 24/2 & SGD          & N/A            & 5-fold CV ensemble \\
 & ViCOROB  & 128$^3$          & 2    & SGD          & A40 / A30      & 2-model ensemble \\
 & IMR      & Auto (nnU-Net)   & 2    & SGD          & RTX 3090       & Sliding window \\
 & CEMRG    & N/A              & N/A  & N/A          & N/A            & Single model \\
 & PEDRETS  & 128$^3$          & 2    & AdamW        & A40            & Single best \\
 & BIGR     & 128$\times$160$\times$128 & 2 & AdamW   & A100           & Sliding window \\
\addlinespace
\midrule
\multirow{8}{*}{2025}
 & MIC      & 192$^3$          & 16   & SGD          & N/A            & Multi-model ensemble \\
 & Eagles   & Auto (nnU-Net)   & 2    & SGD          & N/A            & 5-fold softmax ens. \\
 & WoodyLoveAI & 128$^3$       & 2    & AdamW        & A100           & Sliding window \\
 & PCLab-HIT & 128$^3$         & 2    & SGD / Adam   & RTX 3090       & Sliding + mirroring \\
 & SlicenDice & 128$^3$        & 2    & SGD          & N/A            & 5-fold ensemble \\
 & GenMI    & 128$\times$160$\times$112 & small & Default & N/A        & Sliding + CC filter \\
 & CALADAN  & 128$^3$          & N/A  & AdamW        & N/A            & Single best \\
 & MedAI    & 128$^3$          & 2    & SGD          & A100           & Sliding (step 0.5) \\
\bottomrule
\end{tabular}
\end{table}

\begin{table}[!htbp]
\centering
\caption{\textbf{Task~1 (venous-phase) leaderboard --- Egypt.}
Teams ranked by average rank across all metrics. Distance metrics (RVD, ASSD, RMSD, MSD) in $\times 10^3$\,mm; include 9{,}001\,mm penalty for unmatched lesions. $\uparrow$: higher is better; $\downarrow$: lower is better. Best per column in bold. \textit{Note:} the 9,001 mm penalty for unmatched lesions is necessary to prevent trivial solutions (e.g., predicting nothing, which would otherwise yield artificially optimal surface distances). However, because real lesion surface distances are orders of magnitude smaller than this penalty, the penalty dominates the per-case aggregate for all four distance metrics, causing RVD, ASSD, RMSD, and MSD to converge to effectively identical values that primarily reflect detection completeness rather than boundary accuracy. Readers interested in boundary quality on successfully-detected lesions should focus on Dice, VOE, and the IoU-thresholded precision/recall columns. This behaviour applies to Supplementary Tables~S6--S11.}
\label{supp:tab:t1_egypt}
\footnotesize
\setlength{\tabcolsep}{4pt}
\begin{tabular}{@{}rlccccccccccc@{}}
\toprule
\textbf{Rank} & \textbf{Team} & Dice $\uparrow$ & Dice$_g$ $\uparrow$ & Prec $\uparrow$ & Rec $\uparrow$ & Prec$_{>0}$ $\uparrow$ & Rec$_{>0}$ $\uparrow$ & VOE $\downarrow$ & RVD $\downarrow$ & ASSD $\downarrow$ & RMSD $\downarrow$ & MSD $\downarrow$ \\
\midrule
1 & \textbf{MIC} & \textbf{0.791} & \textbf{0.845} & \textbf{0.573} & 0.381 & \textbf{0.907} & 0.602 & 0.791 & \textbf{6.3} & \textbf{6.3} & \textbf{6.3} & \textbf{6.3} \\
2 & GenBold & 0.644 & 0.697 & 0.494 & 0.394 & 0.817 & 0.650 & \textbf{0.782} & 6.5 & 6.5 & 6.5 & 6.5 \\
3 & WLAI & 0.620 & 0.660 & 0.538 & 0.345 & 0.855 & 0.549 & 0.817 & 6.6 & 6.6 & 6.6 & 6.6 \\
4 & PASTA & 0.700 & 0.807 & 0.404 & 0.345 & 0.839 & 0.717 & 0.845 & 6.9 & 6.9 & 6.9 & 6.9 \\
5 & PCLab-HIT & 0.608 & 0.658 & 0.382 & \textbf{0.438} & 0.687 & \textbf{0.788} & 0.831 & 6.7 & 6.7 & 6.7 & 6.7 \\
6 & Eagles & 0.637 & 0.796 & 0.469 & 0.296 & 0.860 & 0.544 & 0.838 & 7.0 & 7.0 & 7.0 & 7.0 \\
7 & SliceNDice & 0.543 & 0.754 & 0.227 & 0.425 & 0.400 & 0.748 & 0.885 & 7.4 & 7.4 & 7.4 & 7.4 \\
8 & VoxTell & 0.650 & 0.779 & 0.214 & 0.341 & 0.300 & 0.478 & 0.902 & 7.6 & 7.6 & 7.6 & 7.6 \\
9 & Caladan & 0.280 & 0.528 & 0.061 & 0.265 & 0.105 & 0.451 & 0.967 & 8.5 & 8.5 & 8.5 & 8.5 \\
\bottomrule
\end{tabular}
\end{table}

\begin{table}[!htbp]
\centering
\caption{\textbf{Task~1 (venous-phase) leaderboard --- China.}
Teams ranked by average rank across all 11 metrics used in the official TriALS evaluation (see Methods).
Distance metrics (RVD, ASSD, RMSD, MSD) are reported in $\times 10^3$\,mm and include a fixed 9{,}001\,mm penalty for each unmatched lesion (false negative or false positive).
$\uparrow$: higher is better; $\downarrow$: lower is better. Best per column in bold. See Supplementary Table~S6 footnote for interpretation of the distance metric columns
}
\label{supp:tab:t1_china}
\footnotesize
\setlength{\tabcolsep}{4pt}
\begin{tabular}{@{}rlccccccccccc@{}}
\toprule
\textbf{Rank} & \textbf{Team} & Dice $\uparrow$ & Dice$_g$ $\uparrow$ & Prec $\uparrow$ & Rec $\uparrow$ & Prec$_{>0}$ $\uparrow$ & Rec$_{>0}$ $\uparrow$ & VOE $\downarrow$ & RVD $\downarrow$ & ASSD $\downarrow$ & RMSD $\downarrow$ & MSD $\downarrow$ \\
\midrule
1 & \textbf{MIC} & \textbf{0.721} & \textbf{0.833} & \textbf{0.217} & \textbf{0.170} & \textbf{0.217} & 0.170 & \textbf{0.926} & \textbf{8.1} & \textbf{8.1} & \textbf{8.1} & \textbf{8.1} \\
2 & WLAI & 0.557 & 0.758 & 0.185 & 0.136 & 0.215 & 0.159 & 0.942 & 8.2 & 8.2 & 8.2 & 8.2 \\
3 & PCLab-HIT & 0.556 & 0.720 & 0.101 & 0.148 & 0.116 & 0.170 & 0.957 & 8.4 & 8.4 & 8.4 & 8.4 \\
4 & Eagles & 0.542 & 0.704 & 0.122 & 0.102 & 0.176 & 0.148 & 0.961 & 8.5 & 8.5 & 8.5 & 8.5 \\
5 & PASTA & 0.499 & 0.597 & 0.047 & 0.091 & 0.083 & 0.159 & 0.979 & 8.7 & 8.7 & 8.7 & 8.7 \\
6 & SliceNDice & 0.447 & 0.568 & 0.024 & 0.080 & 0.055 & \textbf{0.182} & 0.988 & 8.8 & 8.8 & 8.8 & 8.8 \\
7 & VoxTell & 0.656 & 0.802 & 0.015 & 0.136 & 0.017 & 0.159 & 0.991 & 8.9 & 8.9 & 8.9 & 8.9 \\
8 & GenBold & 0.376 & 0.359 & 0.035 & 0.057 & 0.085 & 0.136 & 0.986 & 8.8 & 8.8 & 8.8 & 8.8 \\
9 & Caladan & 0.425 & 0.759 & 0.022 & 0.091 & 0.035 & 0.148 & 0.988 & 8.8 & 8.8 & 8.8 & 8.8 \\
\bottomrule
\end{tabular}
\end{table}

\begin{table}[!htbp]
\centering
\caption{\textbf{Task~1 (venous-phase) leaderboard --- External.}
Teams ranked by average rank across all metrics. Distance metrics (RVD, ASSD, RMSD, MSD) in $\times 10^3$\,mm; include 9{,}001\,mm penalty for unmatched lesions. $\uparrow$: higher is better; $\downarrow$: lower is better. Best per column in bold. See Supplementary Table~S6 footnote for interpretation of the distance metric columns.}
\label{supp:tab:t1_external}
\footnotesize
\setlength{\tabcolsep}{4pt}
\begin{tabular}{@{}rlccccccccccc@{}}
\toprule
\textbf{Rank} & \textbf{Team} & Dice $\uparrow$ & Dice$_g$ $\uparrow$ & Prec $\uparrow$ & Rec $\uparrow$ & Prec$_{>0}$ $\uparrow$ & Rec$_{>0}$ $\uparrow$ & VOE $\downarrow$ & RVD $\downarrow$ & ASSD $\downarrow$ & RMSD $\downarrow$ & MSD $\downarrow$ \\
\midrule
1 & \textbf{MIC} & \textbf{0.750} & 0.820 & \textbf{0.346} & 0.215 & 0.519 & 0.323 & \textbf{0.890} & \textbf{7.6} & \textbf{7.6} & \textbf{7.6} & \textbf{7.6} \\
2 & Eagles & 0.591 & 0.823 & 0.307 & 0.208 & 0.568 & 0.385 & 0.902 & 7.7 & 7.7 & 7.7 & 7.7 \\
3 & PCLab-HIT & 0.665 & \textbf{0.835} & 0.229 & \textbf{0.231} & 0.427 & \textbf{0.431} & 0.911 & 7.8 & 7.8 & 7.8 & 7.8 \\
4 & WLAI & 0.595 & 0.807 & 0.333 & 0.192 & \textbf{0.573} & 0.331 & 0.902 & 7.8 & 7.8 & 7.8 & 7.8 \\
5 & PASTA & 0.513 & 0.479 & 0.177 & 0.200 & 0.374 & 0.423 & 0.931 & 8.1 & 8.1 & 8.1 & 8.1 \\
6 & GenBold & 0.535 & 0.491 & 0.126 & 0.200 & 0.271 & \textbf{0.431} & 0.944 & 8.2 & 8.2 & 8.2 & 8.2 \\
7 & SliceNDice & 0.535 & 0.643 & 0.089 & 0.185 & 0.189 & 0.392 & 0.959 & 8.4 & 8.4 & 8.4 & 8.4 \\
8 & VoxTell & 0.686 & 0.746 & 0.007 & 0.215 & 0.010 & 0.331 & 0.995 & 8.9 & 8.9 & 8.9 & 8.9 \\
9 & Caladan & 0.464 & 0.781 & 0.042 & 0.154 & 0.077 & 0.285 & 0.976 & 8.7 & 8.7 & 8.7 & 8.7 \\
\bottomrule
\end{tabular}
\end{table}

\begin{table}[!htbp]
\centering
\caption{\textbf{Task~2 (NCCT) leaderboard --- Egypt.}
Visible paradigm includes segmentation and detection metrics; combined paradigm includes detection metrics only.
Distance metrics in $\times 10^3$\,mm. $\uparrow$: higher is better; $\downarrow$: lower is better. Best per column bolded separately within each paradigm. See Supplementary Table~S6 footnote for interpretation of the distance metric columns.}
\label{supp:tab:t2_egypt}
\footnotesize
\setlength{\tabcolsep}{3.5pt}
\begin{tabular}{@{}rllccccccccccc@{}}
\toprule
\textbf{Rank} & \textbf{Team} & \textbf{Paradigm} & Dice $\uparrow$ & Dice$_g$ $\uparrow$ & Prec $\uparrow$ & Rec $\uparrow$ & Prec$_{>0}$ $\uparrow$ & Rec$_{>0}$ $\uparrow$ & VOE $\downarrow$ & RVD $\downarrow$ & ASSD $\downarrow$ & RMSD $\downarrow$ & MSD $\downarrow$ \\
\midrule
1 & \textbf{MIC} & Visible & \textbf{0.552} & \textbf{0.413} & \textbf{0.409} & 0.233 & \textbf{0.796} & 0.454 & \textbf{0.881} & \textbf{7.4} & \textbf{7.4} & \textbf{7.4} & \textbf{7.4} \\
  & & Combined & \textbf{0.471} & \textbf{0.442} & 0.280 & 0.076 & \textbf{0.817} & 0.221 & --- & --- & --- & --- & --- \\
\addlinespace[3pt]
2 & MedAI & Visible & 0.430 & 0.294 & 0.312 & 0.209 & 0.651 & 0.436 & 0.907 & 7.7 & 7.7 & 7.7 & 7.7 \\
  & & Combined & 0.359 & 0.343 & 0.183 & 0.058 & 0.725 & 0.230 & --- & --- & --- & --- & --- \\
\addlinespace[3pt]
3 & PASTA & Visible & 0.353 & 0.324 & 0.271 & 0.215 & 0.620 & \textbf{0.491} & 0.916 & 7.8 & 7.8 & 7.8 & 7.8 \\
  & & Combined & 0.290 & 0.329 & 0.109 & 0.041 & 0.682 & 0.256 & --- & --- & --- & --- & --- \\
\addlinespace[3pt]
4 & Eagles & Visible & 0.350 & 0.324 & 0.222 & \textbf{0.252} & 0.351 & 0.399 & 0.914 & 7.8 & 7.8 & 7.8 & 7.8 \\
  & & Combined & 0.323 & 0.384 & 0.146 & 0.078 & 0.395 & 0.212 & --- & --- & --- & --- & --- \\
\addlinespace[3pt]
5 & WLAI & Visible & 0.230 & 0.272 & 0.299 & 0.178 & 0.485 & 0.288 & 0.919 & 7.9 & 7.9 & 7.9 & 7.9 \\
  & & Combined & 0.230 & 0.272 & \textbf{0.299} & \textbf{0.178} & 0.485 & \textbf{0.288} & --- & --- & --- & --- & --- \\
\addlinespace[3pt]
6 & PCLab-HIT & Visible & 0.374 & 0.257 & 0.234 & 0.196 & 0.504 & 0.423 & 0.918 & 7.9 & 7.9 & 7.9 & 7.9 \\
  & & Combined & 0.264 & 0.295 & 0.161 & 0.064 & 0.526 & 0.209 & --- & --- & --- & --- & --- \\
\addlinespace[3pt]
7 & VoxTell & Visible & 0.435 & 0.382 & 0.116 & 0.196 & 0.209 & 0.356 & 0.947 & 8.3 & 8.3 & 8.3 & 8.3 \\
  & & Combined & 0.330 & 0.367 & 0.083 & 0.067 & 0.227 & 0.183 & --- & --- & --- & --- & --- \\
\addlinespace[3pt]
8 & Caladan & Visible & 0.101 & 0.140 & 0.019 & 0.129 & 0.037 & 0.252 & 0.994 & 8.9 & 8.9 & 8.9 & 8.9 \\
  & & Combined & 0.147 & 0.194 & 0.018 & 0.058 & 0.050 & 0.163 & --- & --- & --- & --- & --- \\
\bottomrule
\end{tabular}
\end{table}

\begin{table}[!htbp]
\centering
\caption{\textbf{Task~2 (NCCT) leaderboard --- China.}
Visible paradigm includes segmentation and detection metrics; combined paradigm includes detection metrics only.
Distance metrics in $\times 10^3$\,mm. $\uparrow$: higher is better; $\downarrow$: lower is better. Best per column bolded separately within each paradigm. See Supplementary Table~S6 footnote for interpretation of the distance metric columns.}
\label{supp:tab:t2_china}
\footnotesize
\setlength{\tabcolsep}{3.5pt}
\begin{tabular}{@{}rllccccccccccc@{}}
\toprule
\textbf{Rank} & \textbf{Team} & \textbf{Paradigm} & Dice $\uparrow$ & Dice$_g$ $\uparrow$ & Prec $\uparrow$ & Rec $\uparrow$ & Prec$_{>0}$ $\uparrow$ & Rec$_{>0}$ $\uparrow$ & VOE $\downarrow$ & RVD $\downarrow$ & ASSD $\downarrow$ & RMSD $\downarrow$ & MSD $\downarrow$ \\
\midrule
1 & \textbf{MIC} & Visible & \textbf{0.621} & 0.599 & 0.135 & \textbf{0.106} & 0.203 & \textbf{0.160} & \textbf{0.953} & \textbf{8.4} & \textbf{8.4} & \textbf{8.4} & \textbf{8.4} \\
  & & Combined & \textbf{0.586} & 0.589 & 0.122 & \textbf{0.047} & 0.189 & \textbf{0.074} & --- & --- & --- & --- & --- \\
\addlinespace[3pt]
2 & WLAI & Visible & 0.488 & \textbf{0.760} & \textbf{0.176} & 0.096 & \textbf{0.216} & 0.117 & 0.954 & \textbf{8.4} & \textbf{8.4} & \textbf{8.4} & \textbf{8.4} \\
  & & Combined & 0.466 & \textbf{0.738} & \textbf{0.157} & 0.042 & \textbf{0.216} & 0.058 & --- & --- & --- & --- & --- \\
\addlinespace[3pt]
3 & Eagles & Visible & 0.513 & 0.635 & 0.101 & 0.085 & 0.177 & 0.149 & 0.966 & 8.6 & 8.6 & 8.6 & 8.6 \\
  & & Combined & 0.502 & 0.639 & 0.101 & 0.042 & 0.165 & 0.068 & --- & --- & --- & --- & --- \\
\addlinespace[3pt]
4 & MedAI & Visible & 0.501 & 0.581 & 0.119 & 0.085 & 0.194 & 0.138 & 0.966 & 8.5 & 8.5 & 8.5 & 8.5 \\
  & & Combined & 0.454 & 0.529 & 0.104 & 0.037 & 0.194 & 0.068 & --- & --- & --- & --- & --- \\
\addlinespace[3pt]
5 & PCLab-HIT & Visible & 0.404 & 0.528 & 0.030 & 0.074 & 0.060 & 0.149 & 0.985 & 8.8 & 8.8 & 8.8 & 8.8 \\
  & & Combined & 0.368 & 0.474 & 0.022 & 0.026 & 0.056 & 0.068 & --- & --- & --- & --- & --- \\
\addlinespace[3pt]
6 & Caladan & Visible & 0.286 & 0.515 & 0.009 & 0.043 & 0.031 & 0.149 & 0.996 & 8.9 & 8.9 & 8.9 & 8.9 \\
  & & Combined & 0.321 & 0.561 & 0.013 & 0.032 & 0.031 & \textbf{0.074} & --- & --- & --- & --- & --- \\
\addlinespace[3pt]
7 & PASTA & Visible & 0.311 & 0.311 & 0.046 & 0.053 & 0.120 & 0.138 & 0.983 & 8.8 & 8.8 & 8.8 & 8.8 \\
  & & Combined & 0.289 & 0.284 & 0.037 & 0.021 & 0.102 & 0.058 & --- & --- & --- & --- & --- \\
\addlinespace[3pt]
8 & VoxTell & Visible & 0.430 & 0.430 & 0.008 & 0.064 & 0.016 & 0.128 & 0.995 & 8.9 & 8.9 & 8.9 & 8.9 \\
  & & Combined & 0.382 & 0.406 & 0.005 & 0.021 & 0.016 & 0.063 & --- & --- & --- & --- & --- \\
\bottomrule
\end{tabular}
\end{table}

\begin{table}[!htbp]
\centering
\caption{\textbf{Task~2 (NCCT) leaderboard --- External.}
Visible paradigm includes segmentation and detection metrics; combined paradigm includes detection metrics only.
Distance metrics in $\times 10^3$\,mm. $\uparrow$: higher is better; $\downarrow$: lower is better. Best per column bolded separately within each paradigm. See Supplementary Table~S6 footnote for interpretation of the distance metric columns.}
\label{supp:tab:t2_external}
\footnotesize
\setlength{\tabcolsep}{3.5pt}
\begin{tabular}{@{}rllccccccccccc@{}}
\toprule
\textbf{Rank} & \textbf{Team} & \textbf{Paradigm} & Dice $\uparrow$ & Dice$_g$ $\uparrow$ & Prec $\uparrow$ & Rec $\uparrow$ & Prec$_{>0}$ $\uparrow$ & Rec$_{>0}$ $\uparrow$ & VOE $\downarrow$ & RVD $\downarrow$ & ASSD $\downarrow$ & RMSD $\downarrow$ & MSD $\downarrow$ \\
\midrule
1 & \textbf{MIC} & Visible & \textbf{0.674} & \textbf{0.840} & \textbf{0.362} & 0.154 & \textbf{0.603} & 0.257 & \textbf{0.912} & \textbf{7.9} & \textbf{7.9} & \textbf{7.9} & \textbf{7.9} \\
  & & Combined & \textbf{0.651} & \textbf{0.791} & \textbf{0.345} & 0.092 & \textbf{0.586} & 0.157 & --- & --- & --- & --- & --- \\
\addlinespace[3pt]
2 & WLAI & Visible & 0.516 & 0.793 & 0.269 & 0.132 & 0.463 & 0.228 & 0.935 & 8.1 & 8.1 & 8.1 & 8.1 \\
  & & Combined & 0.483 & 0.742 & 0.224 & 0.069 & 0.493 & 0.152 & --- & --- & --- & --- & --- \\
\addlinespace[3pt]
3 & Eagles & Visible & 0.537 & 0.761 & 0.226 & 0.140 & 0.417 & 0.257 & 0.938 & 8.2 & 8.2 & 8.2 & 8.2 \\
  & & Combined & 0.539 & 0.729 & 0.179 & 0.069 & 0.429 & 0.166 & --- & --- & --- & --- & --- \\
\addlinespace[3pt]
4 & PCLab-HIT & Visible & 0.563 & 0.756 & 0.172 & 0.162 & 0.305 & 0.287 & 0.939 & 8.2 & 8.2 & 8.2 & 8.2 \\
  & & Combined & 0.541 & 0.707 & 0.133 & 0.078 & 0.305 & 0.180 & --- & --- & --- & --- & --- \\
\addlinespace[3pt]
5 & MedAI & Visible & 0.502 & 0.702 & 0.217 & 0.132 & 0.446 & 0.272 & 0.942 & 8.2 & 8.2 & 8.2 & 8.2 \\
  & & Combined & 0.477 & 0.651 & 0.169 & 0.065 & 0.470 & 0.180 & --- & --- & --- & --- & --- \\
\addlinespace[3pt]
6 & VoxTell & Visible & 0.579 & 0.714 & 0.007 & \textbf{0.191} & 0.012 & \textbf{0.309} & 0.996 & 8.9 & 8.9 & 8.9 & 8.9 \\
  & & Combined & 0.553 & 0.678 & 0.006 & \textbf{0.097} & 0.011 & 0.180 & --- & --- & --- & --- & --- \\
\addlinespace[3pt]
7 & PASTA & Visible & 0.405 & 0.434 & 0.117 & 0.125 & 0.269 & 0.287 & 0.960 & 8.4 & 8.4 & 8.4 & 8.4 \\
  & & Combined & 0.369 & 0.383 & 0.041 & 0.028 & 0.276 & \textbf{0.184} & --- & --- & --- & --- & --- \\
\addlinespace[3pt]
8 & Caladan & Visible & 0.352 & 0.582 & 0.029 & 0.125 & 0.057 & 0.243 & 0.990 & 8.8 & 8.8 & 8.8 & 8.8 \\
  & & Combined & 0.373 & 0.590 & 0.026 & 0.069 & 0.057 & 0.152 & --- & --- & --- & --- & --- \\
\bottomrule
\end{tabular}
\end{table}

\begin{table}[!htbp]
\centering
\caption{\textbf{Task~1 (venous-phase) ranking summary.}
Average rank across all metrics and three test centres. Lower is better.}
\label{supp:tab:t1_rank}
\small
\begin{tabular}{@{}rlcccc@{}}
\toprule
\textbf{Rank} & \textbf{Team} & \textbf{China} & \textbf{Egypt} & \textbf{External} & \textbf{Avg Rank} \\
\midrule
1 & \textbf{MIC} & 1.14 & 1.73 & 2.14 & 1.67 \\
2 & WLAI & 2.68 & 4.05 & 3.59 & 3.44 \\
3 & PCLab-HIT & 3.32 & 4.45 & 3.14 & 3.64 \\
4 & Eagles & 4.50 & 5.27 & 2.82 & 4.20 \\
5 & PASTA & 5.50 & 4.32 & 5.50 & 5.11 \\
6 & GenBold & 7.00 & 3.09 & 5.73 & 5.27 \\
7 & SliceNDice & 6.73 & 6.00 & 6.82 & 6.52 \\
8 & VoxTell & 6.86 & 7.09 & 7.27 & 7.08 \\
9 & Caladan & 7.27 & 9.00 & 8.00 & 8.09 \\
\bottomrule
\end{tabular}
\end{table}

\begin{table}[!htbp]
\centering
\caption{\textbf{Task~2 (NCCT) ranking summary.}
Average rank across combined and visible paradigms and three test centres. Lower is better.}
\label{supp:tab:t2_rank}
\small
\begin{tabular}{@{}rlccccccc@{}}
\toprule
 & & \multicolumn{3}{c}{\textbf{Combined}} & \multicolumn{3}{c}{\textbf{Visible}} & \\
\cmidrule(lr){3-5} \cmidrule(lr){6-8}
\textbf{Rank} & \textbf{Team} & China & Egypt & Ext. & China & Egypt & Ext. & \textbf{Avg Rank} \\
\midrule
1 & \textbf{MIC} & 1.92 & 2.00 & 2.00 & 1.73 & 1.18 & 1.59 & 1.74 \\
2 & WLAI & 2.67 & 3.67 & 3.92 & 2.09 & 5.55 & 3.14 & 3.50 \\
3 & Eagles & 3.08 & 4.00 & 4.00 & 3.50 & 4.18 & 3.50 & 3.71 \\
4 & MedAI & 3.67 & 3.42 & 4.83 & 3.36 & 2.64 & 4.86 & 3.80 \\
5 & PCLab-HIT & 5.67 & 5.17 & 3.83 & 5.55 & 5.32 & 3.77 & 4.88 \\
6 & PASTA & 6.83 & 4.83 & 6.17 & 5.68 & 3.27 & 6.09 & 5.48 \\
7 & VoxTell & 6.92 & 5.17 & 4.50 & 7.27 & 5.86 & 5.91 & 5.94 \\
8 & Caladan & 5.25 & 7.75 & 6.75 & 6.82 & 8.00 & 7.14 & 6.95 \\
\bottomrule
\end{tabular}
\end{table}

\clearpage
\section*{Supplementary Note 1: Algorithmic Descriptions of Participating Methods}

This note summarizes the methodological approach of each participating team across both challenge editions. Teams are grouped by edition and ordered by final ranking.

\subsection*{TriALS 2024 --- Egyptian Test Cohort}

\paragraph{Team MIC (DKFZ, Germany).}
The top-ranked team employed a progressive fine-tuning strategy built on the MultiTalent framework~\cite{ulrich2023multitalent}, using a ResEnc-L U-Net architecture within nnU-Net~\cite{isensee2021nnunet}. Pre-training was conducted on 30 public CT datasets covering diverse anatomical structures. The model was then fine-tuned on a combined liver-specific dataset (LiTS + TriALS multi-phase data; 371 images), followed by task-specific fine-tuning on venous-phase or non-contrast CT data for 200~epochs. Multi-task learning (liver + lesion segmentation) was used throughout. This was the only team to participate in both editions.

\paragraph{Team ViCOROB (Universitat de Girona, Spain).}
This team proposed an ensemble of two nnU-Net models: (1)~a vanilla model trained solely on TriALS data, and (2)~a model pre-trained on LiTS and fine-tuned on TriALS with liver pseudo-masks. Probability maps were combined via arithmetic mean with a threshold of 0.2. For Task~2 (non-contrast), a two-step approach used organizer-provided pseudo-labels as an additional input channel, with a ResidualEncoderUNet augmented by attention gates. Pre-training on all four contrast phases preceded NCCT-specific fine-tuning.

\paragraph{Team IMR (Shanghai Jiao Tong University, China).}
A three-step pipeline was employed: (1)~liver segmentation using nnU-Net trained on four public datasets (SLIVER07, LiTS, CHAOS, AMOS; 470 scans total), (2)~lesion segmentation within the liver VOI using a full-resolution nnU-Net trained from scratch on the 60 TriALS cases, and (3)~post-processing to remove false positives outside the liver mask. Voxels were resampled to 0.8\,mm isotropic.

\paragraph{Team CEMRG (Imperial College London, United Kingdom).}
A two-stage approach combined self-supervised pre-training with a student--teacher contrastive framework, followed by fine-tuning of an xLSTM-UNet model. The xLSTM layers were intended to capture temporal dependencies across contrast phases. No external data was used.

\paragraph{Team PEDRETS (Medical University of Innsbruck, Austria).}
MedNeXt-Large~\cite{roy2023mednext} with 5$\times$5 kernels ($\sim$63M parameters) was first trained on LiTS, then fine-tuned on the challenge dataset for both tasks. Training was conducted on an NVIDIA A40 GPU with AdamW optimizer and Dice + CE loss.

\paragraph{Team BIGR (Erasmus MC, Netherlands).}
A single 3D full-resolution MedNeXt-Large model was trained within the nnU-Net framework from scratch (no pre-training). Task~1 was trained for 100~epochs and Task~2 for 300~epochs on an NVIDIA A100 GPU. The team opted against ensembling due to inference time constraints.

\subsection*{TriALS 2025 --- Multi-Centre Test Cohorts}

\paragraph{Team MIC (DKFZ, Germany).}
The 2025 submission extended the 2024 approach significantly. MultiTalent pre-training was expanded to 86 public datasets (27{,}174 volumes across five modalities) using a ResEnc-L architecture. Fine-tuning incorporated five additional public liver tumor datasets (LiTS, HCC, LiverMets, HCC-TACE, WAW-TACE; 880 patients total) alongside TriALS multi-phase data. A key innovation was restricting training to liver regions of interest (liver bounding box with 5\% margin), ignoring signals outside the liver. Model selection was based on 1{,}000-iteration bootstrapped ranking. Final submissions used ensembles of up to three models from different training configurations.

\paragraph{Team Eagles (NewGiza University, Egypt).}
A 3D nnU-Net with a ResNet medium encoder was used with standard preprocessing (isotropic resampling, intensity normalization, patch-based training). Five-fold cross-validation with softmax ensembling was employed. No external data or pre-training was used.

\paragraph{Team WoodyLoveAI (National Yang Ming Chiao Tung University, Taiwan).}
A custom multi-scale encoder block inspired by ConvNeXt and DeepLabV3+ was integrated into nnU-Net V2. A three-stage transfer learning strategy was used: pre-training on MSD Liver (131 cases), fine-tuning on a Primary Liver Cancer CECT dataset (278 cases), and final fine-tuning on TriALS. The custom architecture ($\sim$31M parameters) consistently outperformed standard nnU-Net and ResEnc variants.

\paragraph{Team PCLab-HIT (Harbin Institute of Technology, China).}
For Task~1, the nnU-Net backbone was augmented with a contrastive learning projection head (InfoNCE loss, temperature 0.2) and hard-sample weighting for improved small-lesion discrimination. For Task~2 (non-contrast), 3D multi-layer decoder affinity constraints were introduced to enhance boundary sensitivity. All models used five-fold cross-validation on a single RTX~3090 GPU.

\paragraph{Team SlicenDice (Universidade do Porto, Portugal).}
A Response Fusion Attention (RFA) mechanism was integrated into a Residual U-Net within the nnU-Net framework. Training combined the TriALS and LiTS datasets (211 volumes). Five-fold cross-validation with 200~epochs was used. This team focused on Task~1 only.

\paragraph{Team GenMI (MBZUAI, United Arab Emirates).}
An nnU-Net-style encoder--decoder with a hybrid Dice--Focal loss was used with comprehensive augmentation (mixup, cutout, elastic deformations, histogram matching). Models were trained for 500~epochs. This team focused on Task~1 only.

\paragraph{Team CALADAN (KNUST, Ghana).}
MedNeXt-M with 3$\times$3$\times$3 kernels was trained using a staged iterative strategy: three rounds of 50-epoch training with checkpoint refinement. For Task 2 (non-contrast), transfer learning from contrast-enhanced phases (arterial + venous as two-channel input) preceded non-contrast fine-tuning.

\paragraph{Team MedAI (Korea University, South Korea).}
STU-Net~\cite{huang2023stunet} with region-based supervision was applied within the nnU-Net~V2 preprocessing framework. Labels were grouped into hierarchical regions (whole liver = liver + lesion; lesion only), and a compound Dice + CE loss was used. Training used all available cases (fold\_all) for 1{,}000~epochs on an NVIDIA A100. This team participated in Task~2 only.
\clearpage

\clearpage

\end{document}